\documentclass{article}

\PassOptionsToPackage{numbers, compress}{natbib}

\usepackage[main, final]{neurips_2026}

\usepackage[utf8]{inputenc} 
\usepackage[T1]{fontenc}    
\usepackage{hyperref}       
\usepackage{url}            
\usepackage{booktabs}       
\usepackage{amsfonts}       
\usepackage{nicefrac}       
\usepackage{microtype}      
\usepackage{xcolor}         
\usepackage{makecell}
\usepackage{amsmath}
\usepackage{multirow}
\usepackage{graphicx}
\usepackage{pifont}   
\usepackage{colortbl} 
\usepackage{xcolor}
\usepackage{subcaption}
\usepackage{wrapfig}
\usepackage[normalem]{ulem}
\usepackage{comment}
\definecolor{revnew}{RGB}{0,70,200}
\definecolor{revold}{RGB}{200,30,30}
\newif\ifshowchanges
\showchangesfalse
\ifshowchanges
  \DeclareRobustCommand{\revadd}[1]{{\color{revnew}#1}}
  \DeclareRobustCommand{\revdel}[1]{{\color{revold}\sout{#1}}}
  \newenvironment{revaddblock}{\color{revnew}}{}
  
\else
  \DeclareRobustCommand{\revadd}[1]{#1}
  \DeclareRobustCommand{\revdel}[1]{}
  \newenvironment{revaddblock}{}{}
  \excludecomment{revdelblock}
\fi
\newcommand{\cmark}{\ding{51}}
\newcommand{\xmark}{\ding{55}}

\title{VACE: Learning Geometrically Structured Representations for Time Series Anomaly Detection}

\author{
Alberto D. Cencillo$^{1\revadd{*}}$, \;
Leonardo Concepción$^{1}$, \;
Isaac Triguero$^{2,1}$, \;
Julián Luengo$^{2,1}$ \\[0.8em]
$^1$Andalusian Research Institute in Data Science and Computational Intelligence (DaSCI)\\
$^2$Department of Computer Science and Artificial Intelligence (DECSAI)\\
University of Granada, Granada, 18071, Spain \\[0.8em]
\revadd{\texttt{\{albertodiazcen, leonar2cp\}@ugr.es}, \texttt{\{triguero, julianlm\}@decsai.ugr.es}}
}

\begin{document}

\maketitle

\begin{abstract}
    Anomaly detection in multivariate time series is a critical task across a wide range of real-world applications, where abnormal behaviour is rare, labels are unavailable, and the cost of a miss is high. The central challenge is learning a characterisation of normality precise enough to flag deviations. Representation self-supervised learning, typically through contrastive approaches, addresses this by embedding temporal patches into a latent space where normality occupies a well-defined region, with anomalies detected by geometric deviation. However, contrastive approaches shape this space indirectly through pair-sampling heuristics, providing no explicit control over the geometric structure on which distance-based scoring operates. That geometry is thus left to chance. We present VACE (\textbf{V}elocity-\textbf{A}ligned \textbf{C}hannel  \textbf{E}mbeddings), a self-supervised anomaly detection method that represents normality as a locally smooth, directionally coherent trajectory in the embedding space. To this end, VACE trains a channel-aware encoder through a velocity-consistency objective, with no negatives and no synthetic anomalies, so that normal trajectories are locally smooth and aligned. At test time, a Mahalanobis positional score and a directional score computed against a bank of normal directions are combined multiplicatively, the directional score acting as a gate that amplifies the positional one where the local dynamics are atypical. Despite its simplicity, VACE achieves state-of-the-art performance on TSB-AD-M under rigorous evaluation, significantly outperforming more complex methods trained on substantially larger budgets.
\end{abstract}

\section{Introduction}

The proliferation of multivariate sensor data in industrial, medical, and infrastructure systems has made Multivariate Time Series Anomaly Detection (MTSAD) a critical component of automated monitoring pipelines. In industrial systems, subtle deviations in sensor readings can indicate abnormal operating regimes \cite{hundman2018detecting}. In server infrastructures, shifts in resource usage often accompany performance degradation. In medical monitoring, irregular physiological patterns may reflect clinically relevant changes in patient state \cite{greenwald1990improved}. Across these domains, normal behaviour is abundant and structured, whereas anomalies are rare, heterogeneous, and costly to miss. This fundamental asymmetry has led to approaches that learn exclusively from normal data, framing anomaly detection as the problem of modelling normality and flagging deviations as anomalies \cite{schmidl2022anomaly}.

In MTSAD, normality is not a fixed, well-defined concept: it depends on the operating regime and evolves over time. A localised deviation in a single channel or a sudden shift in global system dynamics can each be anomalous \cite{blazquez2021review}, yet whether any of these constitutes a true anomaly is determined by what the system considers normal at that moment. Detecting anomalies, therefore, requires a representation of normality that is stable, bounded, and concrete enough to serve as a reference. However, raw input space cannot provide this property, given its inherent variability.

\begin{figure}[t]
  \centering
  \begin{subfigure}[t]{0.48\textwidth}
    \centering
    \includegraphics[width=\linewidth]{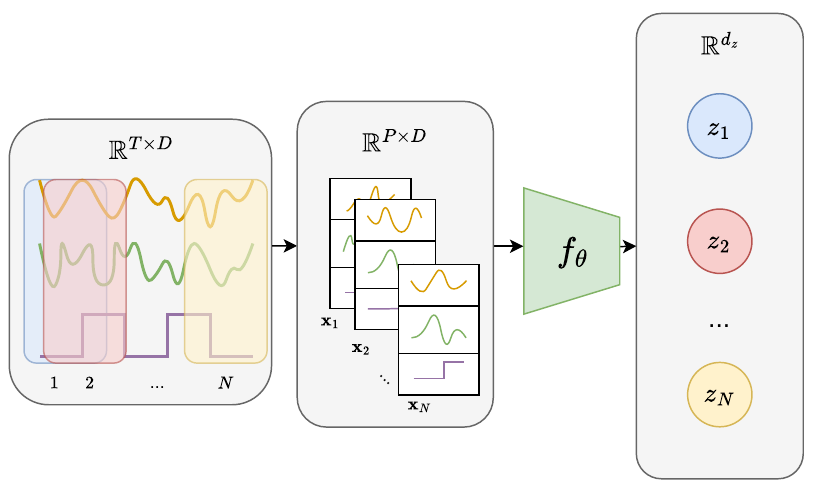}
    \caption{A sliding window extracts $N$ overlapping patches from the multivariate time series $\mathbf{X}$. Each patch $\mathbf{x}_{i}$ is encoded by $f_\theta$ into an embedding vector $z_i$.}
    \label{fig:patches}
  \end{subfigure}\hfill%
  \begin{subfigure}[t]{0.495\textwidth}
    \centering
    \includegraphics[width=\linewidth]{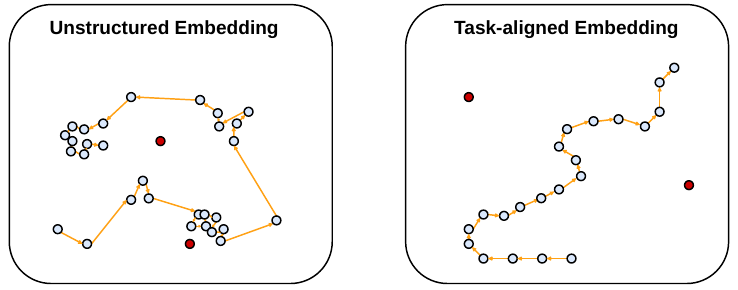}
    \caption{A fragmented, directionally incoherent distribution (left) provides no reliable geometric reference for anomaly scoring (red). A smooth, directionally coherent trajectory (right) makes both positional and directional deviation well-defined signals.}
    \label{fig:embedding_spaces}
  \end{subfigure}
  \caption{Patch embeddings form a trajectory whose geometry determines anomaly detectability.}
  \label{fig:embedding_trajectories}
\end{figure}

Representation learning \cite{bengio2013representation} offers a principled response to this challenge. Originally motivated in computer vision and natural language processing, these methods map raw inputs into an embedding space where the structure implicit in the original data becomes tractable \cite{chen2020simple}. More recently, they have been applied to MTSAD \cite{park2026paano}, where a time series is divided into temporal patches that are mapped into an embedding space, as shown in Figure \ref{fig:patches}, and normality is characterised by the geometry of the resulting distribution. Anomalies are then scored by a notion of distance from the learned normal region \cite{darban2025carla} rather than from a reconstruction target. Since anomaly labels are rarely available in practice, this setting falls naturally within the Self-Supervised Learning (SSL) paradigm, where representations are learned from the structure of the data itself \cite{jing2020self}. This framing is powerful in principle, but existing methods leave two fundamental questions unanswered \cite{wang2020understanding}. First, \emph{how should the encoder be trained} so that the embedding space acquires geometric properties such as smoothness and directional coherence that may benefit a distance-based scorer? Second, \emph{how should position and motion through the embedding space be jointly scored}, given that anomalies can manifest as a point far from the normal distribution, a point moving in an unusual direction, or both?

In practice, representation SSL for MTSAD is dominated by contrastive objectives \cite{yue2022ts2vec}. To answer the first question, these methods reduce the problem to the construction of positive and negative pairs, making the geometry of the embedding space an indirect consequence of sampling heuristics. As a result, pair selection (temporal contrast, synthetic negatives, farthest patches) determines which representations are separated, but provides no explicit control over the geometry of the normal embeddings, that is, how they are distributed in the embedding space and how they move from one patch to the next, that the downstream detection task operates on \cite{saunshi2022understanding}. This introduces three failure modes. On the one hand, treating all non-overlapping normal windows as negatives disperses semantically similar boundary segments, introducing false negatives \cite{yu2024adversarial}. On the other hand, synthetic negatives based on fixed perturbations provide limited boundary coverage and lack difficulty diversity. Finally, without explicit constraints on the extent of the normal region, normal and anomalous embeddings can mix near the decision surface \cite{xie2026caae}. These distortions corrupt the geometry before the scorer even operates.

The answer to the second question has received comparatively little attention. Most methods produce a single scalar distance from the normal region, collapsing the richer structure of the trajectory into a point-wise measure. Anomalies that are positionally plausible but dynamically atypical, or dynamically smooth but spatially displaced, are underscored by any single-signal criterion, as Figure~\ref{fig:embedding_spaces} illustrates: in a fragmented distribution (left), anomalous points (red) blend with normal boundary segments and resist distance-based separation. In a smooth, directionally coherent trajectory (right), both positional and directional deviation become legible signals.

We present \textbf{VACE} (\textbf{V}elocity-\textbf{A}ligned \textbf{C}hannel \textbf{E}mbeddings), a representation SSL method that addresses both questions from a single geometric commitment: the embedding trajectory of a normal time series should be locally smooth and directionally coherent. This commitment drives the method at two levels. First, we embed each multivariate time series through a channel-aware encoder that preserves per-sensor patterns without diluting them across channels, and train it with an objective that directly targets trajectory coherence. In doing so, we intervene directly on the geometry of the embedding space, imposing a property that we expect to benefit a distance-based scorer, rather than relying on instance separation to induce it as contrastive objectives do. Then, we score anomalies with two complementary signals derived from the same geometry: (i) how far an embedding lies from the learned normal distribution, and (ii) how misaligned its local direction of motion is with normal trajectories. The two are combined multiplicatively. The positional score is the primary signal, and the directional score modulates it, \revdel{leaving it almost unchanged where the dynamics are typical and }amplifying it where \revdel{they are}\revadd{the dynamics are} atypical.

These design choices are realised in two concrete contributions:
\begin{enumerate}
\item A velocity-consistency pretext objective ($\mathcal{L}_\mathrm{vel}$) that directly shapes the embedding space with awareness of the downstream scorer, replacing pair sampling with a geometric prior on trajectory smoothness. A channel-aware encoder based on depthwise-separable convolutions ensures that per-channel anomaly signals are preserved before cross-channel mixing, so that the resulting geometry remains legible at the sensor level.

\item A multiplicative state and dynamics scorer that jointly penalises
positional and directional deviation, in which the directional deviation gates the positional one, amplifying it where the local dynamics are atypical.
\end{enumerate}
We evaluate VACE on TSB-AD-M \cite{liu2024elephant}, a large-scale and comprehensive benchmark for MTSAD, where it achieves state-of-the-art performance, with its largest gains on series whose anomalies alter the shape of the signal, such as electrocardiograms, and at a fraction of the computational cost of competing methods. Code is available on: \url{https://github.com/ari-dasci/S-VACE}.

\section{Related Work}

The TSAD literature can be broadly organised into three paradigms according to how normality is represented and deviation is measured \cite{zamanzadeh2024deep}.

\textbf{Forecasting-based methods} define anomalies as deviations between observed values and the predictions of a model trained on normal data. Early approaches used recurrent networks \cite{rumelhart1986parallel} and LSTMs \cite{hochreiter1997long} as one-step predictors, scoring anomalies by prediction error. DeepAnT
\cite{munir2018deepant} extended this to convolutional predictors under the same residual-scoring paradigm. More recently, forecasting foundation models like LagLlama \cite{rasul2023lag}, TimesFM \cite{das2023decoder} or Chronos \cite{ansari2024chronos} have been adapted to TSAD by treating large prediction residuals as anomaly signals. \citet{faber2025xlstmad} apply the xLSTM architecture \cite{beck2024xlstm} under a multi-step forecasting objective. KAN-AD \cite{zhou2024kan} leverages Kolmogorov--Arnold Networks to model the smooth compositional structure of normal time series.

\textbf{Reconstruction-based methods} train a model to compress and reconstruct normal inputs, scoring anomalies by reconstruction error. Autoencoders \cite{audibert2020usad} and variational formulations \cite{su2019robust} established this paradigm; GAN-based variants \cite{zhou2019beatgan} extended it with adversarial training. Transformer architectures introduced association-discrepancy scoring as a proxy for reconstruction fidelity \cite{xu2021anomaly}, and general-purpose models such as TimesNet \cite{wu2022timesnet} and the foundation model MOMENT \cite{goswami2024moment} have since been applied within this framework.

\textbf{Representation-based methods} map the time series into an embedding space and characterise normality geometrically, scoring anomalies by distance from the learned normal region. TS2Vec \cite{yue2022ts2vec} learns general-purpose temporal representations via hierarchical contrastive learning, using non-overlapping segments as implicit negatives. CAE-AD \cite{zhou2022contrastive} combines a contrastive objective with reconstruction at instance and contextual granularity, though inference still relies on reconstruction error. CARLA \cite{darban2025carla} treats synthetically injected anomalies as hard negatives and scores by nearest-neighbour distance in the resulting space. PaAno \cite{park2026paano} adopts a farthest-patch objective to spread the normal distribution and scores via a memory-bank nearest-neighbour criterion. ASTER \cite{hermary2026aster} also places the decision in the representation space, but replaces distance-based scoring with a classifier trained on pseudo-anomalies generated directly in the latent space.

Several recent works have analysed the structural limits of contrastive training in this setting. \citet{yu2024adversarial} show that under normal-only training, treating all non-overlapping windows as negatives disperses semantically similar boundary segments, introducing false negatives that distort the decision boundary. \citet{xie2026caae} identify complementary failure modes: synthetic negatives from fixed perturbation heuristics lack sufficient boundary coverage and difficulty diversity to shape the decision surface, and the absence of explicit constraints on the extent of the normal region allows normal and anomalous embeddings to mix near the boundary.

\section{Method}

\begin{figure}
    \centering
    \includegraphics[width=1.0\linewidth]{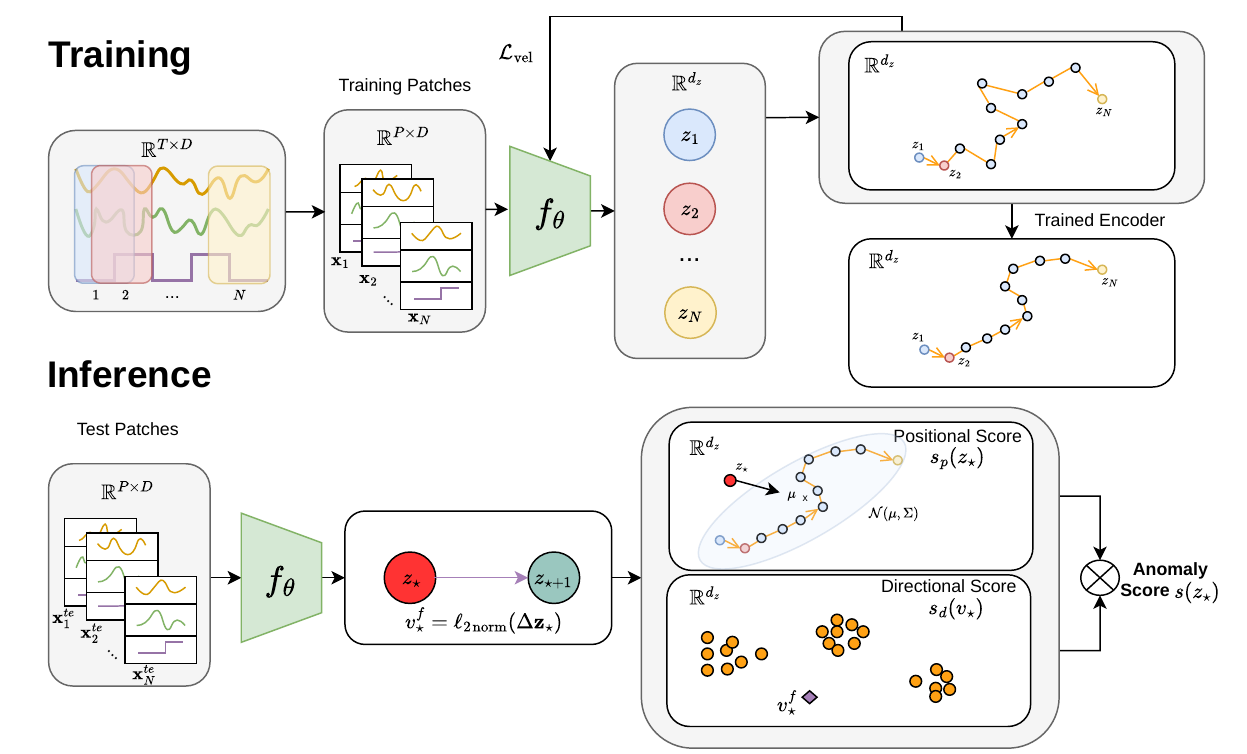}
    \caption{VACE architecture. During training, overlapping patches are encoded by $f_\theta$ under the velocity-consistency objective $\mathcal{L}_{\mathrm{vel}}$, which shapes the normal embedding trajectory $\mathcal{Z}$ into a locally smooth, directionally coherent structure. At inference, test patches are embedded and scored by their positional deviation from $\mathcal{Z}$ and the angular misalignment of their local direction of motion with the normal direction bank.}
    \label{fig:pipeline}
\end{figure}

As illustrated in Figure~\ref{fig:pipeline}, we model a multivariate time series as a trajectory in a learned latent space, where normal behaviour corresponds to smooth and coherent motion. Our approach, VACE, constructs this trajectory by encoding overlapping temporal patches and explicitly shaping their evolution through a velocity-consistency objective. This results in latent representations that are not only spatially structured but also temporally regular, enabling the detection of anomalies as deviations in both position and direction of motion.

\subsection{Geometric Motivation}

Let $\mathbf{X}\in\mathbb{R}^{T\times D}$ denote a multivariate time series of length $T$ with $D$ channels, and let $\mathbf{X}_{a:b}\in\mathbb{R}^{(b-a)\times D}$
denote the contiguous segment spanning timesteps $a,\dots,b-1$. Fixing a patch length $P$, we extract $N = T-P+1$ overlapping patches with unit stride,

\begin{equation}
    \mathbf{x}_i \;=\; \mathbf{X}_{i:\,i+P} \;\in\; \mathbb{R}^{P\times D},
    \qquad i = 1,\dots,N,
\end{equation}

and encode each patch into a vector

\begin{equation}
    \mathbf{z}_i \;=\; f_\theta(\mathbf{x}_i) \;\in\; \mathbb{R}^{d_z},
\end{equation}

where $f_\theta:\mathbb{R}^{P\times D}\!\to\!\mathbb{R}^{d_z}$ is a parametric encoder with learnable parameters $\theta$ and $d_z$ is the embedding dimension, and the ordered tuple
$\mathcal{Z}\,=\,(\mathbf{z}_1,\dots,\mathbf{z}_N)$ is the \emph{normal embedding trajectory} of $\mathbf{X}$, as illustrated in Figure \ref{fig:patches}.

\textbf{Terminology.}
We follow the patch terminology of PatchTST \cite{nie2022time}. A \emph{patch} of length $P$ is the unit encoded by $f_\theta$, and a \emph{window} is the shortest stretch of the series needed for one computation. Training uses windows of $2P$ timesteps, each split into three overlapping patches shifted by $\delta = P/2$ (Section~\ref{sec:pretext}). Scoring uses windows of $1.5P$ timesteps split into two patches, because the second patch is needed to compute the forward direction (Section~\ref{sec:scoring}).

\textbf{Continuity by construction.}
Using unit stride produces maximal overlap between consecutive patches: $\mathbf{x}_i$ and $\mathbf{x}_{i+1}$ share $P-1$ timesteps and differ only at their boundary elements. As a result, when the underlying time series evolves smoothly, consecutive patches are strongly correlated in input space. Since the encoder $f_\theta$ is continuous, this local similarity is preserved in the latent space, leading to embeddings $\mathbf{z}_i = f_\theta(\mathbf{x}_i)$ that vary smoothly across time. In practice, this induces the trajectory $\mathcal{Z}$ that evolves gradually, without abrupt transitions for normal data. Our goal is to exploit this structure by encouraging latent trajectories that remain coherent and evolve smoothly over time, so that deviations from this behaviour can be effectively detected. The training objective shapes only the normal trajectory and applies no explicit separation pressure on anomalies, since that would require committing in advance to how anomalies manifest. Continuity, in turn, constrains the encoder only between nearby inputs. Whether anomalous patches fall off the normal flow is therefore an empirical question, which Appendix~\ref{app:premise} tests directly.

\textbf{Decomposition of the anomaly score.}
Given a test patch embedding $\mathbf{z}_\star$, we define a \emph{positional} score $s_p(\mathbf{z}_\star)$, measuring distance from $\mathbf{z}_\star$ to the training distribution under a covariance-adapted metric, and a \emph{directional} score $s_d(\mathbf{z}_\star)$, measuring how misaligned its local direction of motion is with the bank of normal training directions.  The final score is their multiplicative combination, in which $s_p$ is the primary signal and $s_d$ acts as a gate on it. The encoder is trained with $\mathcal{L}_\mathrm{vel}$ to produce the geometry exploited by both scores.

\subsection{Channel-Aware Patch Encoder}
\label{sec:encoder}
When employing a 1D-CNN for patch encoding, there are two natural variants. A shared-kernel encoder processes all channels jointly from the first layer: cross-channel relationships are preserved, but a single deviating channel can be averaged against the majority of normal ones, potentially masking channel-local anomalies in the embedding. A channel-independent encoder applies separate filters per channel \cite{nie2022time}: per-channel signals are preserved, but inter-channel structure is discarded, making relational anomalies invisible. For the embedding trajectory $\mathcal{Z} $ to be geometrically meaningful, both types of event must remain legible in $\mathbb{R}^{d_z}$, a requirement neither variant alone satisfies. We therefore adopt a two-stage design that separates the two operations: per-channel convolutions first extract local features without cross-channel interference, and a pointwise layer then integrates them into a joint representation.

\begin{figure}[t]
  \centering
  \begin{subfigure}[t]{0.499\textwidth}
    \centering
    \includegraphics[width=\linewidth, height=3.5cm]{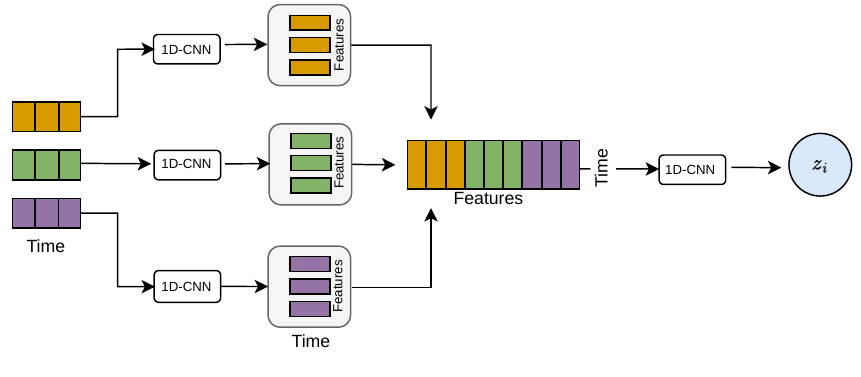}
    \caption{Channel-aware patch encoder.}
    \label{fig:encoder}
  \end{subfigure}\hfill%
  \begin{subfigure}[t]{0.3\textwidth}
    \centering
    \includegraphics[width=\linewidth]{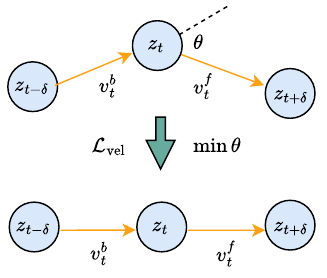}
    \caption{Velocity-consistency loss.}
    \label{fig:velocity_alignment}
  \end{subfigure}
  \caption{Detailed view of two core components: (a) the channel-aware patch encoder and (b) the velocity-consistency loss.}
  \label{fig:miscellanea}
\end{figure}

\textbf{Delayed cross-channel mixing.}
As depicted in Figure \ref{fig:encoder}, we adopt the ``extract per-channel features first, then mix'' principle also
leveraged by ModernTCN \cite{luo2024moderntcn}, and implement it with depthwise-separable convolutions \cite{howard2017mobilenets}. The construction provides two properties simultaneously: (i) a depthwise first stage in which each input channel is processed by its own kernel bank, so per-channel features are captured before any combination takes place and (ii) a pointwise ($1\!\times\!1$) convolution that mixes these enlarged per-channel features into a joint representation. Cross-channel mixing is therefore delayed until per-channel information has already been elevated to a richer feature space.

\textbf{Normalisation.}
Each input patch is first normalised per channel with RevIN \cite{kim2021reversible}, which subtracts its mean and divides it by its standard deviation. The encoder therefore compares the shape of the signal independently of its level and amplitude. Every convolution is followed by batch normalisation (BN) \cite{ioffe2015batch}, and a global average pool over time produces the embedding $\mathbf{z}_i$. We treat BN as part of the architecture, because it keeps the embedding from collapsing under the training objective (Section~\ref{sec:model_analysis}).

\subsection{Velocity-Guided Representation Learning}
\label{sec:pretext}

We train the encoder $f_\theta$ using a self-supervised objective designed to enforce a specific geometric property: the embedding trajectory $\mathcal{Z}$ induced by normal data should be locally coherent, i.e., exhibit consistent direction of motion over time. Our formulation is inspired by RECURVE \cite{shin2024exploiting}, which regularises second-order structure (curvature) for boundary detection. Here, we focus on its first-order counterpart, encouraging local rectilinearity of the trajectory.

\textbf{Velocity-consistency loss.} For a temporal offset $\delta$, we define the backward and forward directions, that is, the $\ell_2$-normalised velocities, at position $t$ as
\begin{equation}
    \mathbf{v}^{b}_t = \frac{\mathbf{z}_t - \mathbf{z}_{t-\delta}}
                            {\operatorname{max}\left(\epsilon, \|\mathbf{z}_t - \mathbf{z}_{t-\delta}\|_2\right)},
    \qquad
    \mathbf{v}^{f}_t = \frac{\mathbf{z}_{t+\delta} - \mathbf{z}_t}          {\operatorname{max}\left(\epsilon,\|\mathbf{z}_{t+\delta} - \mathbf{z}_t\|_2\right)}.
\label{velocity}
\end{equation}

We minimise their cosine misalignment:
\begin{equation}
    \mathcal{L}_\mathrm{vel} \;=\;
    \frac{1}{N} \sum_{t=1}^{N}
    \left(1 - \langle \mathbf{v}^{b}_t, \mathbf{v}^{f}_t \rangle \right).
    \label{eq:velloss}
\end{equation}

This loss enforces that, locally, the trajectory follows a consistent direction (see Figure \ref{fig:velocity_alignment}): the vector entering $\mathbf{z}_t$ is aligned with the one leaving it. Optimum $\mathcal{L}_\mathrm{vel}=0$ is achieved when the trajectory is
straight, since each term $(1 - \langle \mathbf{v}^b_t, \mathbf{v}^f_t \rangle)$ is non-negative  and vanishes if and only if
$\mathbf{v}^b_t = \mathbf{v}^f_t$.
\begin{revaddblock}
The loss is therefore minimised by a globally straight, low-rank trajectory. This degenerate solution is a form of the dimensional collapse described for self-supervised learning \cite{jing2021understanding}. A collapse onto a single point does not minimise the loss, because identical consecutive embeddings give null directions and leave every term at $1$. Three elements of training keep the encoder away from the straight line, namely the short budget of 20 iterations, the linearly decaying loss weight $\lambda_t$ and BN (Section~\ref{sec:encoder}). Appendix~\ref{app:collapse} measures how close training comes to this solution.
\end{revaddblock}


\subsection{Anomaly Scoring via Embedding Geometry}
\label{sec:scoring}

Having trained $f_\theta$ to produce a trajectory $\mathcal{Z}$ that is locally rectilinear on normal data, the scorer reads off the two geometric signatures left by training: how far a test embedding deviates from the normal region, and how misaligned its local direction of motion is with the bank of normal directions.

\textbf{Positional score.}
We fit a Gaussian distribution $\mathcal{N}(\boldsymbol{\mu}, \boldsymbol{\Sigma})$ to the set $\mathcal{Z}$ of all embedded training patches. The positional score of a test embedding $\mathbf{z}_\star$ is defined as the \revdel{squared }Mahalanobis distance \cite{lee2018simple}:
\begin{equation}
    s_p(\mathbf{z}_\star) \;=\; \revadd{\big\|
    (\boldsymbol{\Sigma} + \varepsilon \mathbf{I})^{-1/2}
    (\mathbf{z}_\star - \boldsymbol{\mu})\big\|_2},
\end{equation}
with $\varepsilon = 10^{-6}$ for numerical stability. Mahalanobis reduces to Euclidean distance when $\boldsymbol{\Sigma}\propto\mathbf{I}$ and upweights deviations along low-variance directions otherwise. The regulariser $\varepsilon$ stabilises the inverse when $\boldsymbol{\Sigma}$ is near-singular. This is a common regime for deep embeddings, where a few directions carry most of the variance; $\varepsilon$ prevents a single small eigenvalue from dominating $s_p$.


\textbf{Directional score.}
To characterise the local dynamics of the embedding trajectory, we employ the forward direction $\mathbf{v}_t^f$ with temporal offset $\delta$ defined in Equation \ref{velocity}.

\revdel{We}\revadd{For scoring, we} set $\delta = P/2$, so that the direction captures dynamics over half a patch horizon. Evaluating this over the training sequence yields a set of unit directions $\{\mathbf{v}_t^f\}$, the direction bank. We summarise their distribution by computing $K$ prototypes $\{\mathbf{c}_i\}_{i=1}^K$ via MiniBatch $k$-means on the unit sphere \cite{roth2022towards}.

Given a test embedding $\mathbf{z}_\star$, we compute its associated direction $\mathbf{v}_\star^f$ in the same way. We then identify the $k$ nearest prototypes ($k_{\mathrm{np}}$) to $\mathbf{v}_\star$ in cosine distance. Let $\mathcal{I}_{k_{\mathrm{np}}}(\mathbf{v}_\star^f) \subset \{1,\dots,K\}$ denote the indices of these prototypes. The directional score is defined as:
\begin{equation}
    s_d(\mathbf{z}_\star) \;=\;
    \frac{1}{k_{\mathrm{np}}}
    \sum_{i \in \mathcal{I}_{k_{\mathrm{np}}}(\mathbf{v}_\star^f)}
    \big(1 - \langle \mathbf{v}_\star^f, \mathbf{c}_i \rangle \big).
\end{equation}

By construction, $s_d$ measures the deviation of $\mathbf{v}_\star^f$ from the directions observed during training: normal trajectories produce aligned directions, while anomalous dynamics lead to larger angular deviations.

\textbf{Composition.}
Let $\tilde{s}_p$ and $\tilde{s}_d$ denote \revdel{the normalised versions of $s_p$ and $s_d$, obtained by rescaling to zero mean and unit variance over the test set}\revadd{$s_p$ and $s_d$ after min-max scaling to $[0,1]$ over the test patches of the series}. The final anomaly score is defined as:
\begin{equation}
    s(\mathbf{z}_\star) \;=\; \tilde{s}_p(\mathbf{z}_\star)
    \cdot \big(1 + w \,\tilde{s}_d(\mathbf{z}_\star)\big),
    \qquad w = 1.
\end{equation}

\begin{revaddblock}
The positional score is the primary signal, and the factor $(1 + w\,\tilde{s}_d) \in [1, 1+w]$ modulates it. A patch close to the normal region therefore keeps a low score whatever its direction. An additive combination would instead add the directional score in full, even on series where it carries no information (Appendix~\ref{app:combination}).

\textbf{Boundaries.}
The last $\delta$ patches of a series have no forward neighbour, and we set their directional score to its series mean, so there the final score is proportional to the positional one. Timesteps within $P-1$ steps of either end are covered by fewer patches, and their scores are noisier. Like the baselines of Table~\ref{tab:main}, VACE is an offline detector, since the score of a timestep needs the next $P-1+\delta$ observations.
\end{revaddblock}

\section{Experiments}
\label{sec:experiments}
\subsection{Experimental Setup}

\textbf{Benchmark selection.}
We evaluate on the TSB-AD-M benchmark \cite{liu2024elephant}, a curated collection of 200 multivariate time series designed to address known limitations of prior MTSAD evaluation practices, including label leakage and metric sensitivity to the point-adjustment (PA) heuristic \cite{huet2022local}. The benchmark covers 17 categories spanning NASA telemetry \cite{hundman2018detecting} (MSL, 16 series; SMAP, 27 series), server monitoring \cite{su2019robust} (SMD, 22 series), secure water treatment \cite{mathur2016swat} (SWaT, 2 series), and further domains including medical waveforms, industrial processes, and environmental monitoring. Each series is split into temporally adjacent train and test sets; the training set is anomaly-free, and anomaly labels are only provided for the test set. The 200 time series benchmark is divided into an ``Eval'' set (180 series) and a ``Tuning'' set (20 series) for hyperparameter optimisation.

\textbf{Evaluation metrics.}
We report both threshold-dependent and threshold-free metrics. For threshold-dependent evaluation, we use the point-wise F1 score (F1) and Range-F1 \cite{tatbul2018precision}, the latter weighting full anomaly segments over individual timestamps to better reflect temporal recall. Because thresholding is orthogonal to scoring quality and can mask differences in anomaly ranking \cite{schmidl2022anomaly}, we give primary emphasis to threshold-free metrics: AUC-ROC, AUC-PR, and the Volume Under the Surface variants VUS-ROC and VUS-PR \cite{paparrizos2022volume}. These VUS metrics integrate performance over a range of detection lags and are specifically designed to be robust to the PA inflation problem that affected earlier benchmarks.

\textbf{Compared models and evaluation protocol.}
We compare exclusively against methods that have already been evaluated under the TSB-AD benchmark protocol \cite{liu2024elephant}, so that every reported baseline is obtained under identical conditions. \revadd{Other recent and relevant methods, such as CARLA \cite{darban2025carla} and CAAE \cite{xie2026caae}, do not satisfy this requirement: they were designed for, and evaluated under, window-level protocols and have no published results under TSB-AD. Including them would require adapting parts of their pipelines to the benchmark conditions, so the resulting numbers would reflect our adaptation as much as the methods themselves.} We evaluate against 14 baselines spanning all major paradigms. First, neural network-based methods include TimesNet \cite{wu2022timesnet}, DeepAnT \cite{munir2018deepant}, OmniAnomaly \cite{su2019robust}, USAD \cite{audibert2020usad}, KAN-AD \cite{zhou2024kan}, DADA \cite{shentu2024towards}, and xLSTMAD \cite{faber2025xlstmad}. Second, transformer-based approaches comprise CrossAD \cite{li2025crossad}, AnomalyTransformer \cite{xu2021anomaly}, DCDetector \cite{yang2023dcdetector} and CATCH \cite{wu2025catch}. Third, we consider traditional methods such as PCA \cite{shyu2003novel} and IForest \cite{liu2008isolation}. Finally, we include PaAno \cite{park2026paano} as our primary self-supervised counterpart and the current state-of-the-art on TSB-AD-M. For all baselines, we report results from the standardised TSB-AD-M evaluation protocol \cite{liu2024elephant}.

\textbf{Implementation details.}
The architecture utilises a channel-aware 1D-CNN encoder $f_\theta$ as detailed in Section~\ref{sec:encoder}, employing a four-layer structure with varying kernel sizes and a channel expansion factor of $c_e = 8$. We normalise input patches of length $P=96$ using RevIN \cite{kim2021reversible}, extracting them with unit stride to generate overlapping embeddings. The encoder is trained for 20 iterations under the velocity-consistency objective $\mathcal{L}_\mathrm{vel}$\revadd{, computed with offset $\delta = P$,} with a linearly decaying weight $\lambda_t$, using AdamW with cosine annealing and a batch size of 512. For final scoring, we compute a multiplicative combination ($w=1$) of the Mahalanobis distance and a direction bank, the latter of which is built from up to 500 centroids (10\% of training patches) and evaluated using the three nearest prototypes. Point-level scores are obtained by averaging patch scores over all patches covering each timestep. All experiments were conducted on a single NVIDIA RTX 2080\,Ti. Hyperparameter sensitivity is analysed in Appendix~\ref{sec:sensitivity}. \revadd{VACE is also cheap to run, with lower run times than PaAno and CrossAD on the same hardware (Appendix~\ref{app:compute}).}
\subsection{Comparative Study}

\begin{figure}[t]
  \centering
  \begin{subfigure}[t]{0.38\textwidth}
    \centering
    \includegraphics[width=\linewidth]{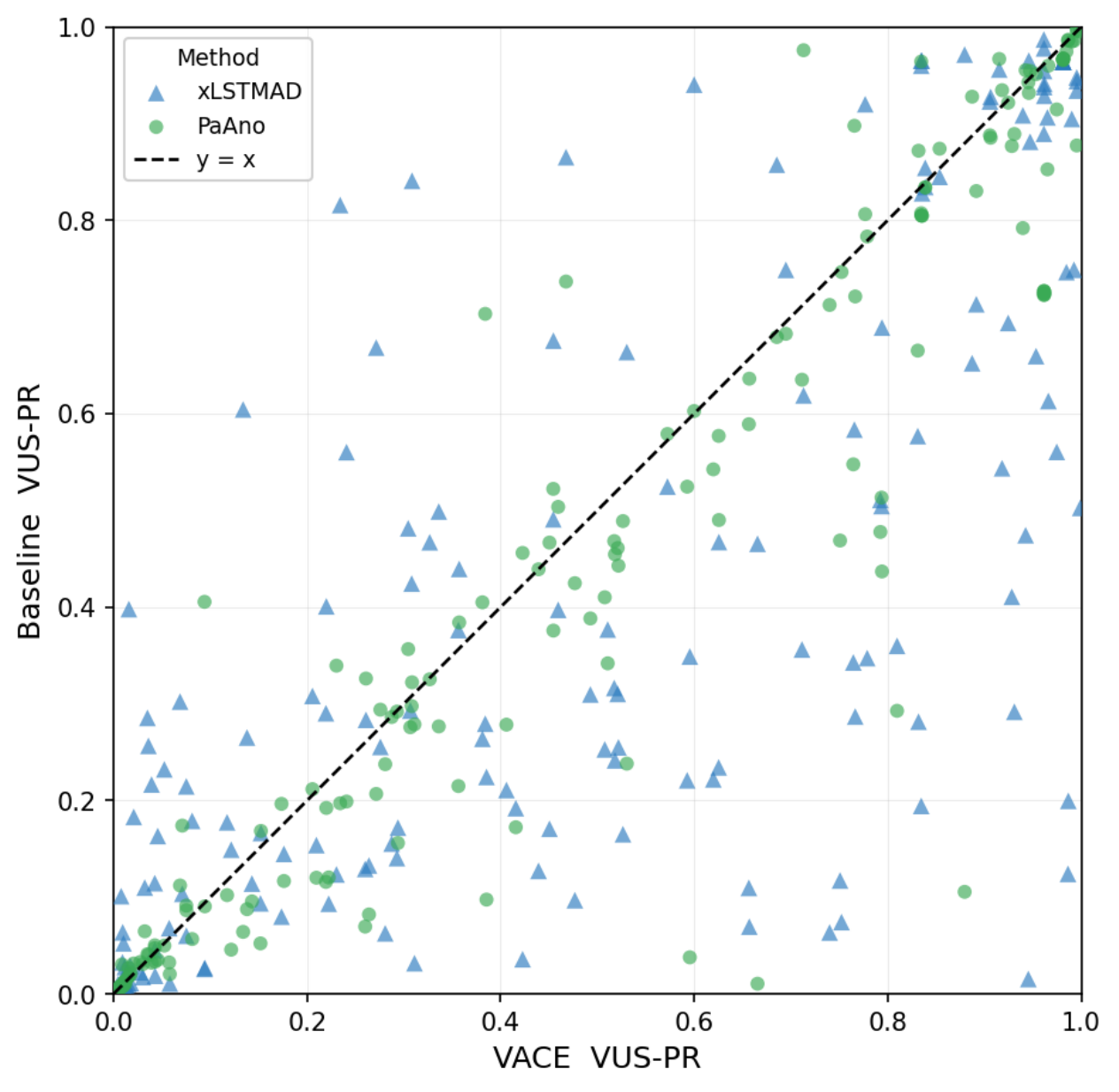}
    \caption{Per-series VUS-PR comparison against VACE. Most points lie below the diagonal. When VACE loses to xLSTMAD-R, the gaps are often larger, indicating an advantage of reconstruction-based methods on some sequence-level anomaly series.}
    \label{fig:scatter}
  \end{subfigure}\hfill%
  \begin{subfigure}[t]{0.60\textwidth}
    \centering
    \includegraphics[width=\linewidth]{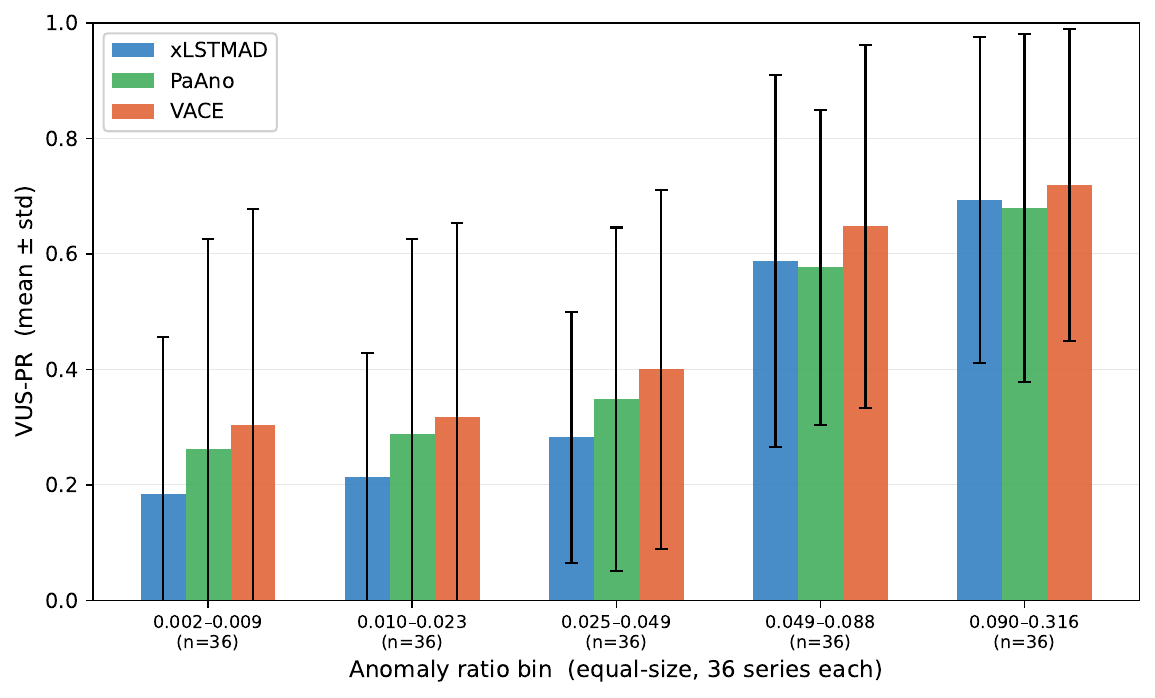}
    \caption{Mean VUS-PR across equal-size anomaly-density bins. \revdel{All methods improve monotonically as anomaly ratio increases. VACE maintains a consistent margin across the full spectrum.} \revadd{All methods score lower as anomalies become scarcer, and VACE ranks first in every bin.}}
    \label{fig:bins}
  \end{subfigure}
  \caption{Per-series and per-density breakdown of VACE against two baselines.}
  \label{fig:digramas}
\end{figure}

\textbf{Main results.} We evaluate VACE on all 180 time series in the TSB-AD-M evaluation set against 14 competitive baselines, as reported in Table~\ref{tab:main}. Our method achieves the highest VUS-PR, AUC-PR, F1, and Range-F1; while VUS-ROC and AUC-ROC are comparable to PaAno, the gains in precision--recall metrics indicate more accurate point-wise detection and more temporally coherent identification of anomalous events. Figure~\ref{fig:scatter} confirms this advantage is broadly distributed: VACE outperforms PaAno on 115 of 180 individual series and xLSTMAD-R on 115 of 180, beating both simultaneously on 79 series against only 29 where both outperform it. Figure~\ref{fig:bins} groups the series into equal-size bins by the proportion of anomalous points in their test split. All methods score lower as anomalies become scarcer, and VACE obtains the highest VUS-PR in every bin. These results substantiate the effectiveness of coupling training and scoring through a shared geometric objective: shaping the embedding trajectory via velocity consistency leads to a representation space in which \revadd{the positional deviation is a strong anomaly signal, which the directional deviation sharpens where the local dynamics are atypical (Section~\ref{sec:model_analysis})}. Additional experiments are reported in Appendix~\ref{sec:additional_experiments}. Notably, VACE reaches this performance with 20 pretext iterations on a single GPU, suggesting that a well-posed geometric objective is a more effective inductive bias than architectural complexity.

\begin{revaddblock}
\textbf{Effect of RevIN.}
The advantage of VACE is largest on series whose anomalies alter the shape of the signal. On the three electrocardiogram (ECG) families, SVDB, MITDB and LTDB (43 of the 180 series), whose anomalies are mostly beats of abnormal form, VACE reaches 0.558 mean VUS-PR, against 0.519 for PaAno and 0.299 for xLSTMAD-R. We attribute much of this advantage to RevIN (Section~\ref{sec:encoder}), which lets the encoder compare waveforms independently of their baseline and amplitude. The same normalisation hides \emph{level-coded} anomalies, which change the level or scale of a channel while the shape of each patch stays plausible (Appendix~\ref{sec:additional_experiments}).
\end{revaddblock}

\begin{table}[t]
  \caption{Average results over 10 seeds on TSB-AD-M. Best result in \textbf{bold}, second best \underline{underlined}. \revadd{For VACE and PaAno, the standard error over the 10 seeds is given in small type. Baseline results are single point estimates from the standardised TSB-AD-M evaluation.}}
  \label{tab:main}
  \centering
  \resizebox{\textwidth}{!}{%
  \begin{tabular}{l l cccccc}
    \toprule
    \textbf{Category} & \textbf{Method}
      & \textbf{VUS-PR} & \textbf{VUS-ROC}
      & \textbf{AUC-ROC} & \textbf{AUC-PR}
      & \textbf{F1} & \textbf{Range-F1} \\
    \midrule
    \multirow{10}{*}{Neural Networks}
      & TimesNet                       & 0.19 & 0.64 & 0.56 & 0.13 & 0.20 & 0.17 \\
      & DeepAnT                            & 0.31 & 0.76 & 0.73 & 0.32 & 0.37 & 0.37 \\
      & OmniAnomaly                    & 0.31 & 0.69 & 0.65 & 0.27 & 0.32 & 0.37 \\
      & USAD                           & 0.30 & 0.68 & 0.64 & 0.26 & 0.31 & 0.37 \\
      & KAN-AD                         & 0.41 & 0.75 & 0.73 & 0.38 & 0.42 & 0.41 \\
      & DADA                           & 0.31 & 0.73 & 0.69 & 0.31 & 0.35 & 0.25 \\
      & xLSTMAD-F (MSE)               & 0.35 & 0.77 & 0.74 & 0.35 & 0.40 & \underline{0.42} \\
      & xLSTMAD-R (MSE)               & 0.37 & 0.72 & 0.68 & 0.32 & 0.38 & 0.36 \\
    \midrule
    \multirow{4}{*}{Transformers}
      & CrossAD                      & 0.33 & 0.77 & 0.74 & 0.34 & 0.38 & 0.37 \\
      & CATCH                   & 0.30 & 0.73 & 0.67 & 0.24 & 0.30 & 0.27 \\
      & AnomalyTransformer           & 0.12 & 0.57 & 0.52 & 0.07 & 0.12 & 0.14 \\
      & DCDetector                   & 0.09 & 0.56 & 0.50 & 0.05 & 0.10 & 0.10 \\
    \midrule
    \multirow{2}{*}{Traditional}
      & PCA         & 0.31 & 0.74 & 0.70 & 0.31 & 0.37 & 0.29 \\
      & IForest     & 0.20 & 0.69 & 0.66 & 0.19 & 0.26 & 0.24 \\
    \midrule
    \multirow{2}{*}{Self-Supervised}
       & PaAno        & \underline{0.43}\revadd{\tiny$\pm$.002} & \textbf{0.79}\revadd{\tiny$\pm$.001} & \textbf{0.76}\revadd{\tiny$\pm$.001} & \underline{0.38}\revadd{\tiny$\pm$.001} & \underline{0.43}\revadd{\tiny$\pm$.001} & 0.41\revadd{\tiny$\pm$.002} \\
     & \cellcolor{gray!12}\textbf{VACE}
     & \cellcolor{gray!12}\textbf{0.48}\revadd{\tiny$\pm$.003}
     & \cellcolor{gray!12}\underline{0.78}\revadd{\tiny$\pm$.001}
     & \cellcolor{gray!12}\underline{0.76}\revadd{\tiny$\pm$.001}
     & \cellcolor{gray!12}\textbf{0.43}\revadd{\tiny$\pm$.003}
     & \cellcolor{gray!12}\textbf{0.46}\revadd{\tiny$\pm$.002}
     & \cellcolor{gray!12}\textbf{0.46}\revadd{\tiny$\pm$.002} \\
    \bottomrule
  \end{tabular}%
  }
\end{table}

\subsection{Model Analysis}
\label{sec:model_analysis}
\begin{table}
  \caption{Ablation study on the 180-series held-out evaluation set of TSB-AD-M
           (mean over 10 seeds).
           \cmark/\xmark\ indicate whether each component is active.
           $\Delta$ is the drop in VUS-PR relative to the full model.}
  \label{tab:ablation}
  \centering
  \resizebox{\textwidth}{!}{
  \begin{tabular}{l cccc | ccc r}
    \toprule
    & \multicolumn{4}{c|}{Component} & \multicolumn{3}{c}{Metric} & \\
    \cmidrule(lr){2-5} \cmidrule(lr){6-8}
    Variant
    & \makecell{Channel\\Encoder}
    & \makecell{Velocity\\Pretext}
    & \makecell{Mahal.\\Scoring}
    & \makecell{Velocity\\Scoring}
    & VUS-PR & AUC-PR & F1 & \multicolumn{1}{c}{$\Delta$VUS-PR} \\
    \midrule
    w/o Channel Encoder  & \xmark & \cmark & \cmark & \cmark & 0.438 & 0.390 & 0.439 & $-$0.039 \\
    w/o Velocity Pretext & \cmark & \xmark & \cmark & \cmark & 0.441 & 0.395 & 0.438 & $-$0.037 \\
    w/o Mahal.\ Scoring  & \cmark & \cmark & \xmark\rlap{$^*$} & \cmark & 0.470 & 0.421 & 0.458 & $-$0.008 \\
    w/o Velocity Scoring & \cmark & \cmark & \cmark & \xmark & 0.472 & 0.423 & 0.459 & $-$0.006 \\
    \midrule
    \rowcolor{gray!12}
    Full model    & \cmark & \cmark & \cmark & \cmark & \textbf{0.478} & \textbf{0.429} & \textbf{0.465} & \multicolumn{1}{c}{---} \\
    \bottomrule
  \end{tabular}
  }
  \vspace{4pt}
  \par
  \raggedright
  {\footnotesize $^*$Mahalanobis distance replaced by a memory-bank scorer.}
\end{table}

\textbf{Ablation study.}
Table~\ref{tab:ablation} reports the contribution of each component, measured as VUS-PR drop relative to the full model averaged over ten random seeds on the 180-series held-out evaluation set. Removing the channel-aware encoder produces the single largest drop of $0.039$ VUS-PR: without depthwise-separable convolutions, per-channel anomaly signals are diluted by early cross-channel mixing before any discriminative feature can be formed. Removing the velocity pretext yields a drop of $0.037$, validating the central design principle that shaping the trajectory geometry during training translates directly into scoring quality. Replacing Mahalanobis scoring with the memory-bank approach used in PaAno~\cite{park2026paano} results in a drop of $0.008$: a single Gaussian fit is sufficient \revadd{on the trained embeddings, where a non-parametric memory bank brings no gain.} Removing velocity scoring leads to a drop of $0.006$. This small individual effect, compared to the pretext gap, \revadd{shows that the benchmark average is carried mostly by the positional score. Appendix~\ref{app:combination} examines what the directional score adds.}

\begin{figure}[t]
  \centering
  \includegraphics[width=0.65\linewidth]{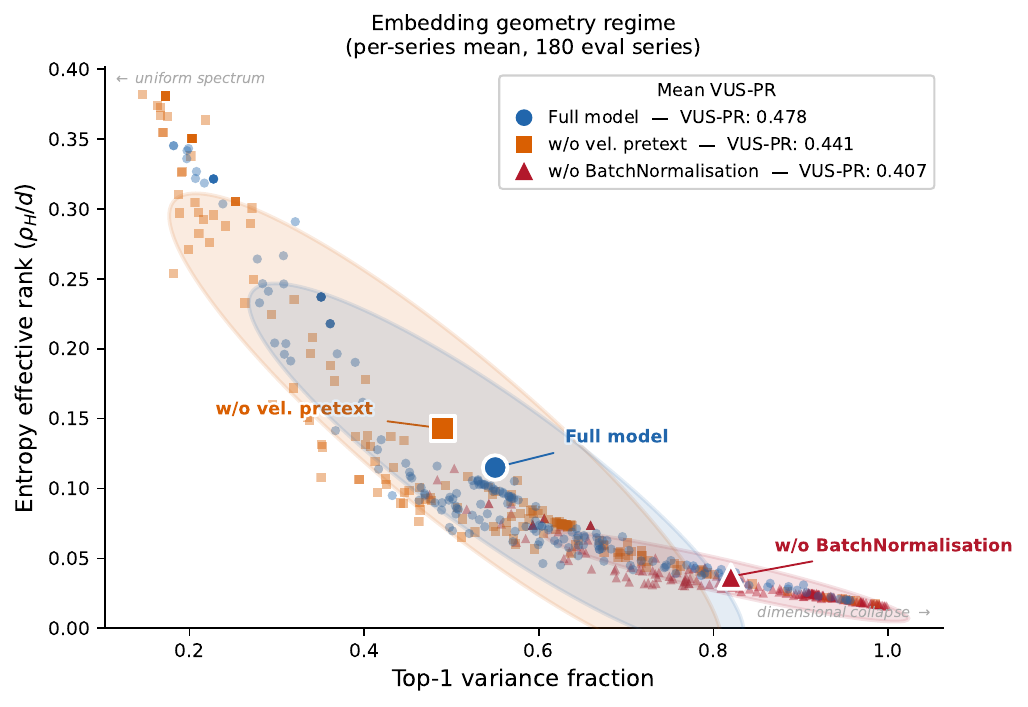}
  \caption{Geometric properties of the embedding distribution (\revadd{metrics defined in Appendix~\ref{sec:geometry_analysis}}) for each ablation configuration, averaged over 10 seeds and 180 evaluation series. The velocity pretext places the full model at moderate anisotropy, the regime in which both scorers are most effective. Without the pretext the distribution is too uniform; without BN it collapses onto a single direction. The mean VUS-PR per configuration confirms that the geometry ordering matches the detection ordering.}
  \label{fig:geometry_regime}
\end{figure}

\textbf{Embedding geometry analysis.}
Figure~\ref{fig:geometry_regime} characterises the embedding distribution under three configurations across 180 series and 10 seeds, using the geometry metrics \revadd{defined in Appendix~\ref{sec:geometry_analysis}, namely two effective ranks of the embedding covariance, the fraction of active dimensions and the fraction of variance carried by the leading direction}. The full model occupies a structured anisotropic regime, with entropy effective rank $0.115/d$, $99.6\%$ of dimensions active, and $55\%$ of variance captured by the leading component. \revadd{This geometry gives both scores a clear structure to exploit:} clear high-variance directions for Mahalanobis distance and dominant motion axes for the direction bank. Removing the velocity pretext pushes the distribution towards uniformity ($\rho_H/d = 0.143$), reducing Mahalanobis to a noisier approximation of isotropic distance. To contrast this with the opposite failure mode, we also evaluate a variant without BN, which is not included as a performance ablation but serves as a geometric diagnostic. The spectrum collapses to $\rho_H/d = 0.037$, the covariance becomes near-singular, and $s_p$ loses sensitivity in all but one direction. The $0.037$ VUS-PR drop from removing the pretext therefore reflects not only lower performance, but the loss of the geometric structure both scorers depend on. Full numerical diagnostics and a per-category breakdown are analysed in Appendix~\ref{sec:geometry_analysis}.

\begin{revaddblock}
\textbf{Batch normalisation.}
A separate single-seed study on 131 of the 200 series follows the geometry of the trajectory during training (Appendix~\ref{app:collapse}). Without BN, the trajectory moves towards the straight, low-rank minimiser of $\mathcal{L}_\mathrm{vel}$ within the 20 deployed iterations, its effective rank halves and the separation of the combined score decreases. BN counteracts this collapse by rescaling every feature at each step, which keeps many latent directions in use. The effect of removing it varies widely between series, since series whose normal embeddings already occupy few dimensions lose little.
\end{revaddblock}

\section{Conclusion}

We present VACE, a self-supervised anomaly detection method built on a single geometric commitment: the embedding trajectory of a normal time series should be
locally smooth and directionally coherent. Its encoder is trained with an objective directly aligned with the downstream scorer, producing the directional structure that positional and velocity scoring rely on. Despite its simplicity, VACE achieves state-of-the-art performance on TSB-AD-M, outperforming methods that are substantially more complex and trained on larger computational budgets, pointing to geometric structure, rather than scale, as the more effective inductive bias. The method can be extended in three places: level-coded anomalies, normal regions without a dominant direction, and the single-Gaussian normality model, which richer densities over the same representation could replace.

\section*{Acknowledgements}
This publication is part of the TSI-100927-2023-1 Project, funded by the Recovery, Transformation and Resilience Plan from the European Union Next Generation through the Ministry for Digital Transformation and the Civil Service, and it is partly funded by the Spanish Ministry of Science and Technology under project PID2023-149128NB-I00, funded by MICIU/AEI/10.13039/501100011033 and by ERDF, EU.

Alberto D. Cencillo holds a DONITOS PhD Scholarship funded by the University of Granada and the particle accelerator IFMIF-DONES.

\bibliographystyle{abbrvnat}
\bibliography{references}

\newpage
\section*{NeurIPS Paper Checklist}


\begin{enumerate}

\item {\bf Claims}
    \item[] Question: Do the main claims made in the abstract and introduction accurately reflect the paper's contributions and scope?
    \item[] Answer: \answerYes{} 
    \item[] Justification: All claims in the abstract and introduction are directly
    supported by the experimental results.

\item {\bf Limitations}
    \item[] Question: Does the paper discuss the limitations of the work performed by the authors?
    \item[] Answer: \answerYes{} 
    \item[] Justification: Limitations are discussed in the Limitations section of the Appendix \ref{app:limitations}.

\item {\bf Theory assumptions and proofs}
    \item[] Question: For each theoretical result, does the paper provide the full set of assumptions and a complete (and correct) proof?
    \item[] Answer: \answerNA{} 
    \item[] Justification: The paper does not include theoretical results.

    \item {\bf Experimental result reproducibility}
    \item[] Question: Does the paper fully disclose all the information needed to reproduce the main experimental results of the paper to the extent that it affects the main claims and/or conclusions of the paper (regardless of whether the code and data are provided or not)?
    \item[] Answer: \answerYes{} 
    \item[] Justification: The architecture, training objective, scoring procedure, and all hyperparameters are fully described in Section \ref{sec:experiments} and the Implementation Details paragraph.

\item {\bf Open access to data and code}
    \item[] Question: Does the paper provide open access to the data and code, with sufficient instructions to faithfully reproduce the main experimental results, as described in supplemental material?
    \item[] Answer: \answerYes{} 
    \item[] Justification: Code is available on \url{https://github.com/ari-dasci/S-VACE}. The benchmark data (TSB-AD-M) is publicly available through its original release.

\item {\bf Experimental setting/details}
    \item[] Question: Does the paper specify all the training and test details (e.g., data splits, hyperparameters, how they were chosen, type of optimizer) necessary to understand the results?
    \item[] Answer: \answerYes{} 
    \item[] Justification: Training and evaluation details are given in the Implementation Details paragraph. Hyperparameter selection is documented in Appendix~\ref{sec:sensitivity}.

\item {\bf Experiment statistical significance}
    \item[] Question: Does the paper report error bars suitably and correctly defined or other appropriate information about the statistical significance of the experiments?
    \item[] Answer: \answerYes{} 
    \item[] Justification: All VACE results are averaged over 10 random seeds.
    The ablation study (Table~\ref{tab:ablation}) reports seed-averaged means,
    and the geometry diagnostics (Table~\ref{tab:geometry}) report mean\,$\pm$\,std
    over 10 seeds. Baseline results are taken from the standardised TSB-AD-M
    evaluation protocol.
    \revadd{Table~\ref{tab:main} reports the standard error of the mean over 10 seeds for VACE and PaAno, the methods we ran ourselves; the remaining baselines are published single point estimates. The score-combination comparison (Table~\ref{tab:combination}) and the premise and collapse studies (Appendices~\ref{app:premise} and~\ref{app:collapse}) report Wilcoxon signed-rank tests over series.}

\item {\bf Experiments compute resources}
    \item[] Question: For each experiment, does the paper provide sufficient information on the computer resources (type of compute workers, memory, time of execution) needed to reproduce the experiments?
    \item[] Answer: \answerYes{} 
    \item[] Justification:  Compute resources are reported in the Implementation Details paragraph, Section \ref{sec:experiments}. \revadd{Run times for training and scoring, parameter counts and their scaling with series length and number of channels are reported in Appendix~\ref{app:compute}.}
    
\item {\bf Code of ethics}
    \item[] Question: Does the research conducted in the paper conform, in every respect, with the NeurIPS Code of Ethics \url{https://neurips.cc/public/EthicsGuidelines}?
    \item[] Answer: \answerYes{} 
    \item[] Justification: The research conforms with the NeurIPS Code of Ethics.

\item {\bf Broader impacts}
    \item[] Question: Does the paper discuss both potential positive societal impacts and negative societal impacts of the work performed?
    \item[] Answer: \answerYes{} 
    \item[] Justification: Broader impacts are discussed in Appendix~\ref{app:broader_impacts}.
    
\item {\bf Safeguards}
    \item[] Question: Does the paper describe safeguards that have been put in place for responsible release of data or models that have a high risk for misuse (e.g., pre-trained language models, image generators, or scraped datasets)?
    \item[] Answer: \answerNA{} 
    \item[] Justification: In this paper, we propose a novel time series Anomaly Detection model based on representation learning. It poses no such risks.

\item {\bf Licenses for existing assets}
    \item[] Question: Are the creators or original owners of assets (e.g., code, data, models), used in the paper, properly credited and are the license and terms of use explicitly mentioned and properly respected?
    \item[] Answer: \answerYes{} 
    \item[] Justification:  All datasets and baselines are properly cited. The TSB-AD-M benchmark is used under its original release terms.

\item {\bf New assets}
    \item[] Question: Are new assets introduced in the paper well documented and is the documentation provided alongside the assets?
    \item[] Answer: \answerYes{} 
    \item[] Justification: The code is released at the anonymous repository linked in the paper.

\item {\bf Crowdsourcing and research with human subjects}
    \item[] Question: For crowdsourcing experiments and research with human subjects, does the paper include the full text of instructions given to participants and screenshots, if applicable, as well as details about compensation (if any)? 
    \item[] Answer: \answerNA{} 
    \item[] Justification: This paper does not involve crowsourcing nor research with human subjects.

\item {\bf Institutional review board (IRB) approvals or equivalent for research with human subjects}
    \item[] Question: Does the paper describe potential risks incurred by study participants, whether such risks were disclosed to the subjects, and whether Institutional Review Board (IRB) approvals (or an equivalent approval/review based on the requirements of your country or institution) were obtained?
    \item[] Answer: \answerNA{} 
    \item[] Justification: This paper does not involve crowsourcing nor research with human subjects.

\item {\bf Declaration of LLM usage}
    \item[] Question: Does the paper describe the usage of LLMs if it is an important, original, or non-standard component of the core methods in this research? Note that if the LLM is used only for writing, editing, or formatting purposes and does \emph{not} impact the core methodology, scientific rigor, or originality of the research, declaration is not required.
    \item[] Answer: \answerNA{} 
    \item[] Justification: No LLMs are used in the core methodology.

\end{enumerate}

\newpage

\appendix

\begin{revaddblock}
\section{Premise Validation}
\label{app:premise}

VACE applies no separation pressure on anomalies, since its training objective shapes only the normal trajectory. Whether anomalous patches nevertheless land away from the normal trajectory is therefore an empirical question, which this appendix tests. On 131 series spanning 15 of the 17 TSB-AD-M categories, we measure how well each score separates anomalous from normal test points with two statistics. The AUC is the probability that an anomalous point is ranked above a normal one, and Cohen's $d$ is the difference between the means of the two populations divided by their pooled standard deviation. Both are computed on the positional score alone ($s_p$) and on the directional score alone ($s_d$).

\paragraph{Positional separation.}
Although nothing in training pushes anomalies away, the anomalous test points lie farther from the normal region than the normal test points on most series. The AUC and Cohen's $d$ of $s_p$ both indicate separation in this direction (AUC $>0.5$, $d>0$) on $75.6\%$ of the series (99 of 131), rising to $93.4\%$ (99 of 106) once the one categorical exception below is excluded.

\paragraph{GHL.}
On GHL, separability is inverted, and anomalous points lie closer to the centre of the normal region than normal test points do. The AUC of $s_p$ is below $0.5$ on $92\%$ of its 25 series, with a mean of $0.28$ and a mean Cohen's $d$ of $-0.75$. GHL is also the hardest category of the benchmark, with no method exceeding $0.083$ VUS-PR (Table~\ref{tab:vuspr_comparison}). It shows that the premise of VACE is falsifiable, and that it fails where anomalies sit in the centre of the normal region.

\paragraph{Directional separation.}
The directional score behaves less uniformly. Its AUC exceeds $0.5$ on $82.4\%$ of the series, with a mean of $0.696$ against $0.766$ for the positional score, and it is markedly category-dependent. It is most informative on the three electrocardiogram (ECG) families, SVDB ($0.90$), MITDB ($0.85$) and LTDB ($0.76$), and on Exathlon ($0.89$). The ECG families share a property that follows from their source. Every heartbeat repeats a similar waveform, so the normal trajectory retraces nearly the same path once per beat and its local direction of motion is highly regular, already at initialisation (Appendix~\ref{app:collapse}). An abnormal beat departs from that path, which the direction bank detects directly. Exathlon, which is not an ECG category, reaches a similar level for a different reason, since its directional separation is mostly produced by training (see below). The directional score is close to chance on GHL ($0.54$), and counter-informative on a few series, such as Daphnet ($0.21$). The multiplicative combination of Section~\ref{sec:scoring} limits the cost of the directional score on such series (Table~\ref{tab:combination}).

\paragraph{Training.}
Training on normal data alone increases the separation of both scores. Between initialisation and the end of the deployed 20-iteration schedule (Wilcoxon signed-rank test over series), the AUC of the positional score improves by $+0.022$ on average (improving on $62.6\%$ of series, $p = 5.7\times10^{-3}$) and that of the directional score by $+0.065$ ($71.8\%$ of series, $p = 4.2\times10^{-7}$). The directional gain exceeds the positional one on $63\%$ of the series ($p = 0.002$), consistent with an objective that constrains the direction of motion of the trajectory rather than its position. It is also concentrated where the trajectory is least regular before training. On the 36 ECG series, whose directions are already well defined at initialisation, the directional AUC moves only from $0.844$ to $0.866$. On the remaining 95 series it rises from $0.550$ to $0.632$ ($+0.081$, $p = 1\times10^{-5}$), while the positional AUC changes by only $+0.012$. Across series, the directional gain decreases with the directional persistence at initialisation (Spearman $\rho = -0.30$, $p = 5\times10^{-4}$), and the largest gains occur on Exathlon ($+0.19$, from $0.70$ to $0.89$), SMAP ($+0.16$) and MSL ($+0.08$). Training therefore sharpens mainly the directional signal, and does so where the normal dynamics are least regular to begin with.

\end{revaddblock}

\begin{revaddblock}
\section{Representational Collapse}
\label{app:collapse}

Section~\ref{sec:pretext} notes that $\mathcal{L}_\mathrm{vel}$ admits a degenerate global minimiser, a globally straight, low-rank trajectory. This appendix measures how close training comes to it. We instrumented the training loop and measured the geometry of the latent trajectory along training on 131 of the 200 TSB-AD-M series, covering 15 of the 17 categories. We compare three regimes, namely the deployed configuration (20 iterations with BN), extended schedules of up to 320 iterations, and a variant without BN.

\paragraph{Point collapse.}
The embedding cloud never contracts, at any schedule length and with or without BN. Over the deployed schedule, its median total variance grows from $3.6\times10^{-5}$ to $3.3\times10^{-2}$, and the median number of active dimensions from 62 to 64 of 64. This agrees with Section~\ref{sec:pretext}, where we noted that a point collapse does not minimise the loss.

\paragraph{Line collapse.}
The straight line, in contrast, is the global minimiser. To detect it at every temporal scale, we use the \emph{directional persistence} at lag $k$,
\begin{equation}
\mathrm{persist}(k) = \mathbb{E}_t\!\left[\cos\!\left(\mathbf{z}_{t+k} - \mathbf{z}_t,\; \mathbf{z}_{t+2k} - \mathbf{z}_{t+k}\right)\right],
\end{equation}
the average cosine between consecutive displacements of the trajectory. Values near $+1$ at every lag at once correspond to the straight-line limit, and negative values mean that the trajectory turns back on itself, as a periodic trajectory does. Table~\ref{tab:persistence} reports it together with the effective rank, the top-1 variance fraction (Appendix~\ref{sec:geometry_analysis}) and the AUC of the directional score.

\begin{table}[h]
  \ifshowchanges\color{revnew}\fi
  \caption{Directional persistence at four lags, effective rank (of 64), top-1 variance fraction and AUC of the directional score, paired over the 131 series. $\delta = 48$.}
  \label{tab:persistence}
  \centering
  \small
  \begin{tabular}{l cccc ccc}
    \toprule
    Configuration & lag 1 & lag 48 & lag 96 & lag 192 & Eff.\ rank & Top-1 var. & AUC $s_d$ \\
    \midrule
    Initialisation (iter.\ 0)        & $+0.31$ & $-0.23$ & $-0.33$ & $-0.45$ & 3.8 & 0.61 & 0.63 \\
    \textbf{BN, 20 iter.\ (deployed)} & $+0.27$ & $-0.22$ & $-0.17$ & $-0.48$ & 3.6 & 0.57 & \textbf{0.70} \\
    No BN, 20 iter.                  & $+0.36$ & $-0.12$ & $+0.03$ & $-0.46$ & 1.8 & 0.79 & 0.63 \\
    BN, 320 iter.                    & $+0.24$ & $+0.08$ & $+0.46$ & $-0.05$ & 1.6 & 0.79 & 0.65 \\
    \bottomrule
  \end{tabular}
\end{table}

Straightening does occur, but only in two situations that the deployed configuration avoids. Without BN it appears within the deployed schedule. In 20 iterations, persistence rises at lags 48 and 96, the effective rank halves and the directional signal degrades across series (Wilcoxon, $p = 7\times10^{-23}$). With BN it requires much longer training. The straightening then starts at the scale on which the loss acts (lag 96, twice the shift $\delta$) and spreads outward. Under the deployed configuration the profile is almost unchanged. The only visible straightening is at the scale of the loss itself (lag 96), and the long-range recurrence at lag 192 is untouched, so the global structure of the trajectory, including its cyclic patterns, is preserved.

\begin{table}[h]
  \ifshowchanges\color{revnew}\fi
  \caption{Directional persistence per category at lags 1, 48, 96 and 192 under the four configurations of Table~\ref{tab:persistence}. Means over the series of each category ($n$). The last row is the pooled mean of Table~\ref{tab:persistence}.}
  \label{tab:persistence_category}
  \centering
  \scriptsize
  \setlength{\tabcolsep}{1.9pt}
  \begin{tabular}{l r cccc cccc cccc cccc}
    \toprule
    & & \multicolumn{4}{c}{Initialisation} & \multicolumn{4}{c}{BN, 20 iter.\ (deployed)} & \multicolumn{4}{c}{No BN, 20 iter.} & \multicolumn{4}{c}{BN, 320 iter.} \\
    \cmidrule(lr){3-6} \cmidrule(lr){7-10} \cmidrule(lr){11-14} \cmidrule(lr){15-18}
    Category & $n$ & 1 & 48 & 96 & 192 & 1 & 48 & 96 & 192 & 1 & 48 & 96 & 192 & 1 & 48 & 96 & 192 \\
    \midrule
    GHL         & 25 & $+.16$ & $-.19$ & $-.21$ & $-.34$ & $+.18$ & $-.26$ & $-.18$ & $-.35$ & $+.17$ & $-.13$ & $-.06$ & $-.31$ & $+.24$ & $-.03$ & $+.05$ & $-.28$ \\
    SVDB        & 21 & $+.71$ & $-.46$ & $-.33$ & $-.52$ & $+.70$ & $-.53$ & $-.11$ & $-.54$ & $+.73$ & $-.57$ & $+.07$ & $-.49$ & $+.53$ & $-.56$ & $+.52$ & $-.16$ \\
    Exathlon    & 17 & $\phantom{+}.00$ & $-.24$ & $-.45$ & $-.54$ & $-.08$ & $-.18$ & $-.31$ & $-.59$ & $+.05$ & $-.03$ & $-.19$ & $-.67$ & $-.03$ & $+.32$ & $+.49$ & $-.20$ \\
    SMAP        & 17 & $+.22$ & $-.12$ & $-.22$ & $-.40$ & $+.15$ & $-.03$ & $-.06$ & $-.42$ & $+.27$ & $-.03$ & $+.13$ & $-.38$ & $+.16$ & $+.25$ & $+.52$ & $+.06$ \\
    SMD         & 16 & $+.02$ & $-.18$ & $-.37$ & $-.44$ & $-.03$ & $-.09$ & $-.10$ & $-.40$ & $+.20$ & $+.18$ & $+.26$ & $-.32$ & $+.07$ & $+.62$ & $+.79$ & $+.37$ \\
    MSL         & 12 & $+.30$ & $+.04$ & $-.34$ & $-.55$ & $+.23$ & $+.04$ & $-.10$ & $-.59$ & $+.36$ & $+.28$ & $+.26$ & $-.57$ & $+.22$ & $+.51$ & $+.84$ & $+.56$ \\
    MITDB       & 10 & $+.74$ & $-.27$ & $-.42$ & $-.48$ & $+.72$ & $-.24$ & $-.33$ & $-.56$ & $+.74$ & $-.12$ & $-.19$ & $-.51$ & $+.53$ & $-.29$ & $+.03$ & $-.60$ \\
    LTDB        &  5 & $+.77$ & $-.46$ & $-.41$ & $-.28$ & $+.74$ & $-.46$ & $-.23$ & $-.43$ & $+.78$ & $-.53$ & $+.13$ & $-.52$ & $+.36$ & $-.27$ & $+.65$ & $+.21$ \\
    SWaT        &  2 & $+.58$ & $-.34$ & $-.43$ & $-.43$ & $+.42$ & $-.35$ & $-.29$ & $-.46$ & $+.65$ & $-.31$ & $-.08$ & $-.43$ & $+.33$ & $+.34$ & $+.60$ & $+.09$ \\
    CATSv2      &  1 & $+.83$ & $-.39$ & $-.49$ & $-.46$ & $+.72$ & $-.42$ & $-.44$ & $-.48$ & $+.82$ & $-.44$ & $-.33$ & $-.49$ & $+.62$ & $-.12$ & $+.26$ & $-.28$ \\
    Daphnet     &  1 & $-.02$ & $-.17$ & $-.46$ & $-.46$ & $-.08$ & $-.15$ & $-.27$ & $-.53$ & $+.02$ & $-.10$ & $-.11$ & $-.57$ & $-.12$ & $+.49$ & $+.62$ & $-.11$ \\
    GECCO       &  1 & $+.15$ & $-.31$ & $-.42$ & $-.48$ & $+.06$ & $-.32$ & $-.35$ & $-.50$ & $+.35$ & $-.21$ & $-.14$ & $-.42$ & $-.09$ & $-.06$ & $+.40$ & $-.32$ \\
    Genesis     &  1 & $+.45$ & $-.10$ & $-.50$ & $-.60$ & $+.54$ & $+.08$ & $-.06$ & $-.80$ & $+.56$ & $+.32$ & $\phantom{+}.00$ & $-.85$ & $+.54$ & $-.48$ & $+.44$ & $-.18$ \\
    OPPORTUNITY &  1 & $+.09$ & $-.32$ & $-.46$ & $-.49$ & $-.05$ & $-.25$ & $-.09$ & $-.54$ & $+.13$ & $-.18$ & $+.19$ & $-.52$ & $-.03$ & $+.56$ & $+.73$ & $+.11$ \\
    PSM         &  1 & $+.14$ & $-.33$ & $-.45$ & $-.47$ & $-.02$ & $-.32$ & $-.40$ & $-.49$ & $+.21$ & $-.27$ & $-.27$ & $-.52$ & $+.10$ & $+.12$ & $+.22$ & $-.54$ \\
    \midrule
    All         & 131 & $+.31$ & $-.23$ & $-.33$ & $-.45$ & $+.27$ & $-.22$ & $-.17$ & $-.48$ & $+.36$ & $-.12$ & $+.03$ & $-.46$ & $+.24$ & $+.08$ & $+.46$ & $-.05$ \\
    \bottomrule
  \end{tabular}
\end{table}

\paragraph{Categories.}
Table~\ref{tab:persistence_category} shows that the approach to the straight line depends on the data. On MSL and SMD, long training makes persistence positive at every lag, including 192, so their trajectories come close to a straight line. These categories contain many constant sensors (a median of $81\%$ of the channels on MSL and $21\%$ on SMD), and their normal embeddings occupy few dimensions from the start. On the ECG families, persistence at lag 192 stays negative even after 320 iterations, and becomes more negative on MITDB. Each heartbeat brings the input back to a similar waveform, so the trajectory keeps returning to the same region, which a straight line cannot do. GHL barely changes under any configuration.

\paragraph{Training length.}
Table~\ref{tab:iterations} retrains VACE on the same 131 series with schedules from 10 to 320 iterations, decaying $\lambda_t$ proportionally so that the schedules are comparable. Every row is a single run under the same seed.

\begin{table}[h]
  \ifshowchanges\color{revnew}\fi
  \caption{Training length, geometry and detection on 131 series (single run, same seed). Eff.\ rank of 64 dimensions.}
  \label{tab:iterations}
  \centering
  \small
  \begin{tabular}{l cccccc}
    \toprule
    Iterations & VUS-PR & Eff.\ rank & Top-1 var. & AUC $s_p$ & AUC $s_d$ & Cohen's $d$ ($s_d$) \\
    \midrule
    10                     & 0.459 & 4.44 & 0.50 & 0.759 & 0.688 & 0.85 \\
    \textbf{20 (deployed)} & \textbf{0.468} & 3.63 & 0.57 & \textbf{0.766} & \textbf{0.696} & \textbf{0.87} \\
    40                     & 0.439 & 2.25 & 0.69 & 0.755 & 0.660 & 0.70 \\
    80                     & 0.429 & 1.80 & 0.75 & 0.751 & 0.637 & 0.58 \\
    160                    & 0.422 & 1.66 & 0.78 & 0.740 & 0.641 & 0.61 \\
    320                    & 0.432 & 1.64 & 0.79 & 0.751 & 0.646 & 0.57 \\
    \bottomrule
  \end{tabular}
\end{table}

Longer schedules lower the effective rank and raise the share of variance carried by the leading direction, and detection quality deteriorates with them. Mean VUS-PR peaks under the deployed schedule. The directional score, which depends on the full local geometry, is the most affected, while the positional score barely moves.

Whether a series degrades depends on how much structure its input carries. The 22 series that improve under longer schedules (mean $+0.24$ VUS-PR) have a median of $22\%$ constant sensors, while the 51 that degrade (mean $-0.17$) have none. For redundant inputs, a low-rank representation is close to the true structure of the data, so collapsing is not necessarily harmful. The ECG categories are the clearest case of the opposite. At initialisation, their trajectories already show a directional persistence of 0.75 at lag 1, against 0.13 for the rest of the benchmark. The objective is therefore nearly satisfied from the start, and further iterations mostly push towards collapse. Of their 36 series, 31 degrade, the worst by around 0.3 VUS-PR. The short schedule thus protects performance where the method is most effective. The degradation comes from the objective itself, and the short schedule, the decaying $\lambda_t$ and BN keep training away from its degenerate optimum.

\end{revaddblock}

\begin{revaddblock}
\section{Score Combination}
\label{app:combination}
\end{revaddblock}

\begin{revaddblock}
Table~\ref{tab:combination} isolates the contribution of the directional score. Each trained model is scored again, without retraining, with the positional score alone, the directional score alone, the product of Section~\ref{sec:scoring} and three additive combinations $\tilde{s}_p + w\,\tilde{s}_d$. The product improves on the positional score alone by a small margin that is consistent across series. It matches the additive rule with a small directional weight and outperforms it as the weight grows. The additive rule adds the directional score in full on every series, including those where it is uninformative, and loses up to $0.28$ VUS-PR on some categories. With the product, the directional score can at most double the positional score, so test points close to the normal region keep a low score, and the worst category loses only $0.07$. The two scores also rank test points differently, with a median absolute Spearman correlation of $0.23$ within a series, so the directional score is not a rescaled copy of the positional one. Its average effect on VUS-PR is small, and it is most informative on a few categories (Appendix~\ref{app:premise}).
\end{revaddblock}

\begin{table}[t]
\begin{revaddblock}
  \caption{Score combination rules on the 180-series evaluation set. Each trained model is re-scored, unchanged, under six rules, so differences are due to the combination rule alone. Single run per series, which places the levels slightly below the 10-seed mean of Table~\ref{tab:main}. $p$: Wilcoxon signed-rank test on the per-series differences with the multiplicative rule.}
  \label{tab:combination}
  \centering
  \begin{tabular}{l cccccc}
    \toprule
    Rule & $\tilde{s}_p$ only & $\tilde{s}_d$ only & \makecell{Multiplicative\\(paper)} & \makecell{Additive\\$w{=}0.5$} & \makecell{Additive\\$w{=}1$} & \makecell{Additive\\$w{=}2$} \\
    \midrule
    Mean VUS-PR & 0.463 & 0.318 & \textbf{0.467} & 0.466 & 0.446 & 0.401 \\
    $p$ vs.\ multiplicative & 0.018 & $10^{-9}$ & --- & 0.440 & 0.130 & 0.001 \\
    \bottomrule
  \end{tabular}
\end{revaddblock}
\end{table}

\section{Sensitivity Analysis}
\label{sec:sensitivity}

We assess the sensitivity of VACE to its five main hyperparameters: patch size $P$, velocity offset $\delta$\revadd{ of the directional score}, number of pretext iterations, velocity weight $w$, and channel expansion factor $c_e$.
All sweeps are conducted on the 20-series tuning subset of TSB-AD-M (disjoint from the 180-series held-out evaluation set used in the main results). Results are stable across a wide range of values; the main sensitivity is to $P$ (small patches hurt substantially) and to the pretext iteration count (both too few and too many degrade performance, with a clear optimum at 20 steps). As depicted in Table \ref{tab:sensitivity}, default values are highlighted.

\begin{table}[h]
\caption{Hyperparameter sensitivity (tuning set, 20 series). 
Each cell shows VUS-PR value. The default configuration is highlighted.\revadd{ Each row is a separate sweep with the remaining hyperparameters at their defaults, so the value of the default configuration varies slightly between rows (0.573 to 0.577).}}
\centering
\setlength{\tabcolsep}{8pt}
\renewcommand{\arraystretch}{1.2}
\begin{tabular}{l ccc}
\toprule
\textbf{Hyperparameter} & \textbf{Low} & \textbf{Default} & \textbf{High} \\
\midrule
Patch size $P$           
& 48 \, (0.464) 
& \cellcolor{gray!15} \textbf{96 \, (0.577)} 
& 128 \, (0.578) \\

Velocity offset $\delta$ 
& 24 \, (0.568) 
& \cellcolor{gray!15} \textbf{48 \, (0.573)} 
& 96 \, (0.553) \\

Iterations            
& 10 \, (0.515) 
& \cellcolor{gray!15} \textbf{20 \, (0.574)} 
& 40 \, (0.542) \\

Velocity weight $w$      
& 0.5 \, (0.574) 
& \cellcolor{gray!15} \textbf{1.0 \, (0.574)} 
& 2.0 \, (0.573) \\

Channel expansion $c_e$  
& 4 \, (0.568) 
& \cellcolor{gray!15} \textbf{8 \, (0.573)} 
& 16 \, (0.542) \\
\bottomrule
\end{tabular}
\label{tab:sensitivity}
\end{table}

\paragraph{Patch size.}
Performance degrades sharply for $P < 96$: small patches carry insufficient temporal context for the velocity-consistency objective to shape a coherent trajectory. Larger patches ($P=128$) perform comparably on the tuning set but reduce the number of patches available for fitting the Mahalanobis distribution.

\paragraph{Velocity offset.}
$\delta = 48$ (half the patch horizon) and $\delta = 24$ (quarter-horizon) yield similar VUS-PR. $\delta = 96$ (full patch horizon) degrades slightly, as the direction vectors span the full patch and become less sensitive to local dynamics.

\paragraph{Pretext iterations.}
Ten pretext steps are insufficient to shape the trajectory geometry, while \revdel{30}\revadd{40} or more \revadd{move the encoder towards the degenerate global minimiser of $\mathcal{L}_\mathrm{vel}$ (Section~\ref{sec:pretext}). Since there are no labels to fit, this degradation is a form of representational collapse. Appendix~\ref{app:collapse} studies this phenomenon.}

\paragraph{Velocity weight.}
VUS-PR is nearly flat across $w \in \{ 0.5, 1.0, 2.0\}$, varying by less than $0.003$.
The directional score contributes a small but consistent improvement over the positional score alone (confirmed by the ablation in Table~\ref{tab:ablation}), and its magnitude is not sensitive to the exact weighting. \revadd{This flatness follows from the multiplicative form, in which the directional score rescales the positional score by a factor between $1$ and $1+w$, so a patch close to the normal region keeps a low score for any weight. An additive combination, in contrast, degrades steadily as the directional weight grows (Table~\ref{tab:combination}).}

\paragraph{Channel expansion.}
$c_e = 8$ and $c_e = 4$ perform similarly; $c_e = 16$ drops notably, consistent with over-parameterisation at the depthwise stage.

\begin{revaddblock}
\section{Computational Cost}
\label{app:compute}

A run-time comparison is only possible for methods run on the same hardware. We therefore report timings for VACE, for PaAno, our direct self-supervised counterpart, and for CrossAD \cite{li2025crossad}, a recent transformer-based detector. All three were run end to end on the same GPU over the 180 evaluation series, and the totals exclude metric computation.

\begin{table}[h]
  \ifshowchanges\color{revnew}\fi
  \caption{Parameters and run time per series (training and scoring), 180 evaluation series, same GPU.}
  \label{tab:runtime}
  \centering
  \small
  \begin{tabular}{l ccc}
    \toprule
    Model & Params (M) & Mean (s) & Median (s) \\
    \midrule
    \textbf{VACE} & 0.31 & \textbf{9.90} & \textbf{7.84} \\
    PaAno         & 0.39 & 23.31 & 8.45 \\
    CrossAD       & 1.33 & 484.63 & 201.24 \\
    \bottomrule
  \end{tabular}
\end{table}

VACE is the fastest of the three in both mean and median, while obtaining the best VUS-PR. The parameter counts show that the difference comes from the training protocol. VACE and PaAno have comparable sizes, and VACE reaches its operating point in 20 gradient steps.

\paragraph{Cost per phase.}
Over the same 180 series, training takes 2.47\,s on average (median 1.87\,s) and test-time scoring 7.42\,s (median 5.58\,s). Training covers patch extraction, the encoder fit under $\mathcal{L}_\mathrm{vel}$ and the model build. Scoring covers embedding the test patches, fitting the Gaussian, building the direction bank, computing both scores and distributing patch scores to points. Training is the smaller part of the budget because of the 20-iteration schedule, so the cost of VACE is dominated by a single pass over the test data.

\paragraph{Scaling.}
The dominant factor in the total cost is the length of the series (Table~\ref{tab:runtime_length}). The number of channels acts on the cost of a single patch, and its effect is cleanest on the training phase, whose volume is fixed by the schedule at 20 iterations of 512 patches (Table~\ref{tab:runtime_channels}). Training time grows with the channel count, but the total does not follow that ordering on this benchmark, because its high-dimensional series are roughly fifteen times shorter than its two-channel recordings, so that sequence length dominates the overall cost.

\begin{table}[h]
  \ifshowchanges\color{revnew}\fi
  \centering
  \small
  \begin{minipage}[t]{0.35\textwidth}
    \centering
    \captionof{table}{Total run time of VACE by series length (train + test).}
    \label{tab:runtime_length}
    \begin{tabular}{l cc}
      \toprule
      Length & Series & Median (s) \\
      \midrule
      $<$20k      & 56 & 2.1 \\
      20k--100k   & 53 & 6.4 \\
      100k--300k  & 63 & 17.5 \\
      $>$300k     &  8 & 30.0 \\
      \bottomrule
    \end{tabular}
  \end{minipage}\hfill
  \begin{minipage}[t]{0.62\textwidth}
    \centering
    \captionof{table}{Median series length and run time of VACE by number of channels.}
    \label{tab:runtime_channels}
    \setlength{\tabcolsep}{4pt}
    \begin{tabular}{l ccccc}
      \toprule
      Channels & Series & Length & Train (s) & Test (s) & Total (s) \\
      \midrule
      2--5   & 54 & 230k & 1.36 & 8.99 & 10.62 \\
      6--30  & 76 & 47k  & 1.68 & 6.05 & 8.04 \\
      31--60 & 42 & 24k  & 1.71 & 1.53 & 3.44 \\
      $>$60  &  8 & 16k  & 5.02 & 3.53 & 9.00 \\
      \bottomrule
    \end{tabular}
  \end{minipage}
\end{table}
\end{revaddblock}

\section{Additional Experiments}
\label{sec:additional_experiments}

Table~\ref{tab:vuspr_comparison} reports per-category VUS-PR for VACE and two competitive baselines.

\begin{table}[t]
  \caption{Per-category VUS-PR comparison. Best result per column in \textbf{bold}.
           VACE is evaluated on the 180-series held-out set (20 tuning series excluded);
           baseline results are from the TSB-AD-M benchmark~\cite{liu2024elephant}.}
  \label{tab:vuspr_comparison}
  \centering
  \resizebox{\textwidth}{!}{
  \begin{tabular}{l ccccccccccccccccc c}
    \toprule
    & \rotatebox{60}{Exathlon}
    & \rotatebox{60}{TAO}
    & \rotatebox{60}{LTDB}
    & \rotatebox{60}{SVDB}
    & \rotatebox{60}{SMAP}
    & \rotatebox{60}{SMD}
    & \rotatebox{60}{GECCO}
    & \rotatebox{60}{MITDB}
    & \rotatebox{60}{MSL}
    & \rotatebox{60}{SWaT}
    & \rotatebox{60}{Genesis}
    & \rotatebox{60}{Daphnet}
    & \rotatebox{60}{PSM}
    & \rotatebox{60}{OPPORTUNITY}
    & \rotatebox{60}{CreditCard}
    & \rotatebox{60}{CATSv2}
    & \rotatebox{60}{GHL}
    & \rotatebox{60}{Overall} \\
    $n$ & 25 & 11 & 4 & 28 & 25 & 20 & 1 & 11 & 14 & 2 & 1 & 1 & 1 & 7 & 1 & 5 & 23 & 180 \\
    \midrule
    VACE      & $0.928$ & $0.692$ & $0.621$ & $\mathbf{0.612}$ & $\mathbf{0.586}$ & $\mathbf{0.434}$ & $\mathbf{0.416}$ & $\mathbf{0.396}$ & $\mathbf{0.344}$ & $0.273$ & $0.271$ & $0.261$ & $0.220$ & $0.108$ & $0.057$ & $0.051$ & $0.010$ & $\mathbf{0.478}$ \\
    PaAno     & $0.814$ & $0.727$ & $\mathbf{0.622}$ & $0.561$ & $0.522$ & $0.358$ & $0.187$ & $0.376$ & $0.245$ & $0.246$ & $0.333$ & $0.274$ & $0.202$ & $0.161$ & $0.027$ & $0.074$ & $\mathbf{0.083}$ & $0.430$ \\
    xLSTMAD-R (MSE)  & $\mathbf{0.944}$ & $\mathbf{0.737}$ & $0.421$ & $0.343$ & $0.275$ & $0.402$ & $0.192$ & $0.141$ & $0.332$ & $\mathbf{0.521}$ & $\mathbf{0.668}$ & $\mathbf{0.283}$ & $\mathbf{0.401}$ & $\mathbf{0.314}$ & $\mathbf{0.068}$ & $\mathbf{0.235}$ & $0.019$ & $0.392$ \\
    \bottomrule
  \end{tabular}
  }
\end{table}

\paragraph{Overall pattern.}
VACE leads on \revdel{7} \revadd{6} of 17 categories.
The two largest categories in the evaluation set, SMAP and SVDB, together account for 53 of 180 series, and VACE leads on both; this concentration in large-$n$ categories drives the overall benchmark gap over PaAno despite trailing on several smaller ones.

\paragraph{Channel-local fault datasets.}
Two of the categories where VACE improves most over PaAno are MSL \revadd{($+0.099$ VUS-PR)} and SMD~\revadd{($+0.076$)}, both of which contain multivariate sensor recordings where faults commonly affect individual channels. This is consistent with the channel-encoder ablation (Table~\ref{tab:ablation}), which shows a drop of $0.039$ VUS-PR when the depthwise-separable encoder is replaced with a standard shared-kernel CNN. The ablation provides direct experimental support for the conclusion that per-channel feature preservation is especially valuable in these two categories; we are cautious about generalising this claim to others where we do not have the same controlled evidence.

\paragraph{Hard categories.}
GHL is the hardest category across all evaluated methods: no model in
Table~\ref{tab:vuspr_comparison} exceeds $0.083$ VUS-PR across its 23 series, and VACE scores the lowest. \revadd{GHL is also the one category on which the premise of VACE fails, since its anomalous points lie closer to the centre of the normal region than the normal test points themselves (Appendix~\ref{app:premise}).}

\paragraph{Comparison with xLSTMAD-R.}
xLSTMAD-R leads on 9 of the 17 categories, most clearly on Genesis, SWaT, OPPORTUNITY, PSM and CATSv2. These are categories where reconstruction-based methods appear to provide a stronger signal. \revadd{The two analyses below identify two properties of these series that account for part of this advantage, without excluding others such as the autoregressive structure or the longer effective context of xLSTMAD.}

\begin{revaddblock}
\paragraph{Failure cases.}
The 29 series on which both PaAno and xLSTMAD-R outperform VACE (Figure~\ref{fig:scatter}) are clustered. Sixteen of them come from GHL (7 of its 23 evaluation series), CATSv2 (5 of 5) and OPPORTUNITY (4 of 7), which together hold 35 of the 180 series. Counting is misleading for GHL, where all three methods of Table~\ref{tab:vuspr_comparison} score below 0.09 VUS-PR. Measured by the VUS-PR actually conceded, the picture is dominated by OPPORTUNITY and CATSv2, which account for half of the deficit over the 29 series, and in both the leading method is the reconstruction-based one.

\paragraph{Embedding concentration.}
These series share a property that can be measured before training and without labels. Their normal region does not concentrate in the embedding space. We quantify it at initialisation as the top-1 variance fraction of the training embeddings (Appendix~\ref{sec:geometry_analysis}), that is, the share of embedding variance carried by the leading direction. The categories where VACE trails hold the least concentrated normal regions of the benchmark (CATSv2 0.24, OPPORTUNITY 0.20, PSM 0.19, SWaT 0.35, with effective ranks of 6 to 12 of the 64 dimensions), and the categories where it leads hold the most concentrated (SMAP 0.82, SVDB 0.73, MITDB 0.66, LTDB 0.65, all with effective ranks below 2.3). 

The diagnostic locates the difficulty in the scorer. Where the normal region has no dominant direction, the single Gaussian that the positional score fits to it offers little contrast, and the separation between normal and anomalous test points shrinks accordingly. On CATSv2, the AUC of the positional score, the probability that an anomalous point is ranked above a normal one, is 0.62, and Cohen's $d$, the standardised difference between the means of the two populations, is 0.31. On the ECG families, the same quantities are 0.82 to 0.92 and 1.5 to 2.5. This is the single-Gaussian limitation discussed in Appendix~\ref{app:limitations}.

\paragraph{Use cases.}
The same diagnostic indicates when VACE is most useful. On the 114 evaluation series whose normal region is concentrated (top-1 variance fraction of at least 0.5 at initialisation), VACE reaches 0.451 mean VUS-PR against 0.300 for xLSTMAD-R, a margin of $+0.151$. On the remaining 66 series the margin is $-0.027$. The clearest cases are SMAP and the ECG families. On SVDB, MITDB and LTDB, VACE reaches 0.558 mean VUS-PR, against 0.519 for PaAno and 0.299 for xLSTMAD-R. Within the representation-based family, the margin of VACE over PaAno is largest where anomalies are confined to one or a few channels, as in MSL and SMD (above). These relations hold over the whole evaluation set and are not explained by dimensionality, but the correlations are moderate, and further experiments are needed before this explanation can be considered complete.

\paragraph{RevIN ablation.}
Table~\ref{tab:revin} complements the diagnostic with the input side of the encoder. We call an anomaly \emph{level-coded} when it is no longer visible once each patch is normalised by its own mean and standard deviation, as RevIN does. Examples are a sustained offset, a slow drift, or a channel whose amplitude leaves its normal range while the shape of the patch remains plausible. A \emph{shape-coded} anomaly, by contrast, changes the form of the signal within the patch, as an abnormal heartbeat does. Without RevIN, each channel of the series is standardised with the mean and standard deviation of its training split. Removing RevIN from VACE costs $0.21$ mean VUS-PR on the ECG families, where anomalies are shape-coded, and gains $0.023$ on the remaining series, with the largest gains on SWaT, Exathlon, MSL, CATSv2 and OPPORTUNITY. RevIN is thus one reason for the strength of VACE on waveform data and one reason for its weakness on part of the categories where reconstruction-based methods lead. Removing RevIN does not help on GHL, whose difficulty comes from the inverted premise described in Appendix~\ref{app:premise}.

\begin{table}[h]
  \ifshowchanges\color{revnew}\fi
  \caption{Mean VUS-PR of VACE with and without RevIN per category, single run per series on the 180 evaluation series (so the levels differ slightly from the 10-seed means of Table~\ref{tab:vuspr_comparison}). $\Delta$: without minus with RevIN. Sorted by $\Delta$.}
  \label{tab:revin}
  \centering
  \small
  \begin{tabular}{l c ccc}
    \toprule
    Category & $n$ & With RevIN & Without RevIN & $\Delta$ \\
    \midrule
    GECCO       &  1 & 0.353 & 0.041 & $-0.312$ \\
    LTDB        &  4 & 0.631 & 0.394 & $-0.237$ \\
    MITDB       & 11 & 0.415 & 0.180 & $-0.235$ \\
    SVDB        & 28 & 0.622 & 0.425 & $-0.197$ \\
    SMD         & 20 & 0.415 & 0.323 & $-0.093$ \\
    PSM         &  1 & 0.218 & 0.182 & $-0.036$ \\
    SMAP        & 25 & 0.596 & 0.579 & $-0.017$ \\
    Daphnet     &  1 & 0.229 & 0.226 & $-0.003$ \\
    CreditCard  &  1 & 0.061 & 0.062 & $+0.001$ \\
    GHL         & 23 & 0.010 & 0.016 & $+0.005$ \\
    TAO         & 11 & 0.695 & 0.714 & $+0.019$ \\
    OPPORTUNITY &  7 & 0.096 & 0.162 & $+0.066$ \\
    CATSv2      &  5 & 0.047 & 0.123 & $+0.076$ \\
    MSL         & 14 & 0.400 & 0.482 & $+0.082$ \\
    Exathlon    & 25 & 0.811 & 0.913 & $+0.102$ \\
    SWaT        &  2 & 0.262 & 0.387 & $+0.125$ \\
    Genesis     &  1 & 0.316 & 0.967 & $+0.651$ \\
    \midrule
    ECG (SVDB, MITDB, LTDB) & 43 & 0.570 & 0.360 & $-0.210$ \\
    Other categories        & 137 & 0.435 & 0.458 & $+0.023$ \\
    All                     & 180 & 0.467 & 0.434 & $-0.033$ \\
    \bottomrule
  \end{tabular}
\end{table}
\end{revaddblock}

\paragraph{Comparison with TSPulse.}
TSPulse~\cite{ekambaramtspulse} is a state-of-the-art time series pretrained model, representing a larger resource regime than VACE. We evaluate it independently using the official implementation from the TSB-AD repository~\cite{liu2024elephant}, under the same protocol applied to all other methods. Due to memory overflow, the OPPORTUNITY category (7 series) could not be evaluated; results are therefore reported over 173 of the 180 evaluation series. Table~\ref{tab:tspulse} reports both the fine-tuned (FT) and zero-shot (ZS) variants. The scores obtained under this protocol are higher than those reported in the original TSPulse paper, likely reflecting differences in benchmark composition and evaluation setup; we report our reproduced numbers for consistency with the TSB-AD-M evaluation framework used throughout this work. Under this protocol, VACE outperforms both TSPulse variants on VUS-PR.

\begin{table}[h]
  \caption{Results on 173 TSB-AD-M evaluation series (OPPORTUNITY excluded). Reproduced from the official TSB-AD implementation.}
  \label{tab:tspulse}
  \centering
  \begin{tabular}{l cccccc}
    \toprule
    Variant & VUS-PR & VUS-ROC & AUC-ROC & AUC-PR & F1 & Range-F1 \\
    \midrule
    TSPulse (ZS) & 0.40 & 0.72 & 0.67 & 0.35 & 0.40 & 0.42 \\
    TSPulse (FT) & 0.44 & 0.76    & 0.72 &  0.38 & 0.43 & 0.42 \\
    \midrule
    \rowcolor{gray!12}
    VACE         & \textbf{0.49} & \textbf{0.80} & \textbf{0.77} & \textbf{0.44} & \textbf{0.48} & \textbf{0.46} \\
    \bottomrule
  \end{tabular}
\end{table}


\section{Geometry Analysis}
\label{sec:geometry_analysis}

\begin{figure}
    \centering
    \includegraphics[width=1\linewidth]{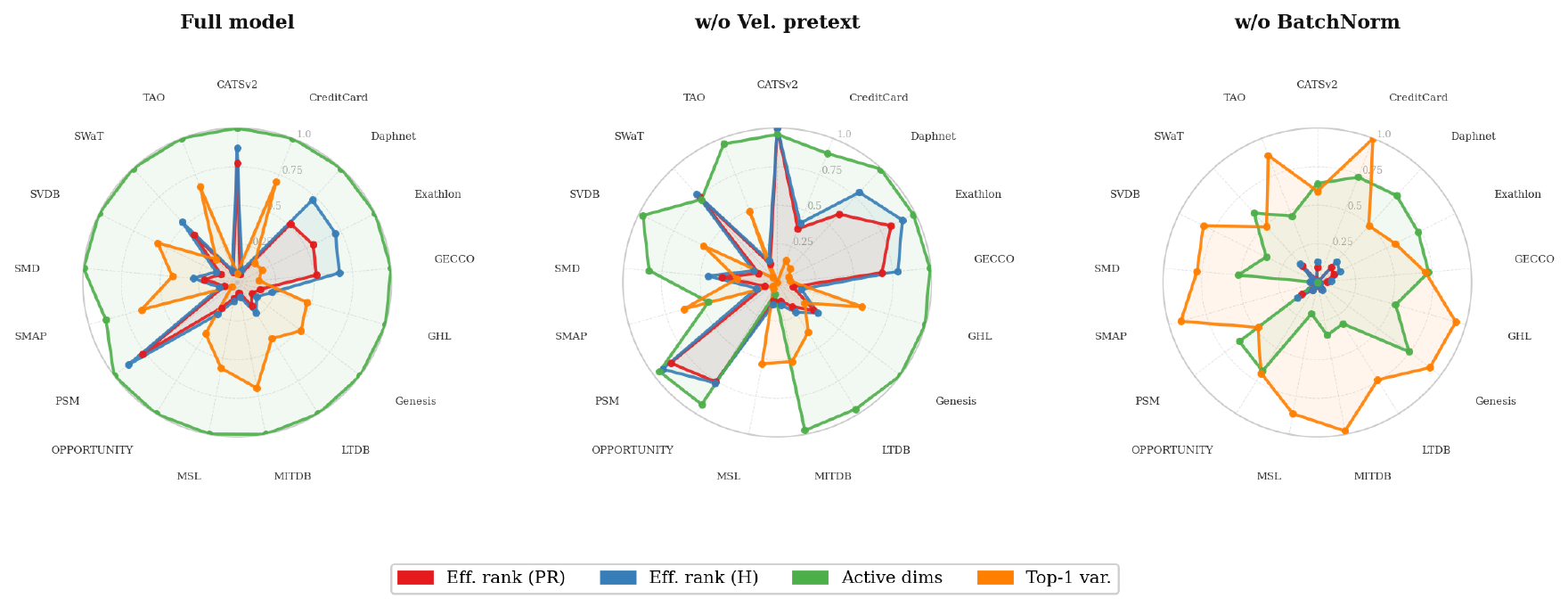}
    \caption{Per-category geometry diagnostics for the full model,
         without the velocity pretext, and without BN.
         The three configurations maintain distinct geometric regimes
         consistently across all 17 dataset families.}
    \label{fig:spider}
\end{figure}

\begin{revaddblock}
\paragraph{Geometry metrics.} \ifshowchanges\emph{(moved from Section~\ref{sec:pretext})}\fi
To characterise the embedding geometry, we report four diagnostics.
The \emph{participation-ratio effective rank} and \emph{entropy effective rank} \cite{roy2007effective} are
\begin{equation}
\rho_\mathrm{PR} =
\frac{\left( \sum_i \lambda_i \right)^2}{\sum_i \lambda_i^2}
\qquad\qquad
\rho_H =
\exp\!\left(-\sum_i p_i \log p_i\right),
\quad
p_i = \frac{\lambda_i}{\sum_j \lambda_j}
\end{equation}
where $\lambda_1 \ge \cdots \ge \lambda_{d_z}$ are the eigenvalues of the sample covariance $\boldsymbol{\Sigma}$ of the embeddings.
Higher values indicate a more isotropic distribution. We also report the \emph{active dimension fraction}, i.e.\ the share of components with explained variance above $1/d_z$, and the \emph{top-1 variance fraction}, defined as $\lambda_1 / \sum_i \lambda_i$. These metrics are commonly used to assess representation quality \cite{garrido2023rankme}. Here they serve to describe the geometry seen by the scorer and to detect collapse \cite{jing2021understanding}.

\end{revaddblock}
This section reports the embedding geometry diagnostics in full.
Table~\ref{tab:geometry} provides the aggregate metrics for the three
configurations (mean $\pm$ std over 10 seeds, 180 evaluation series), complementing
the scatter in Figure~\ref{fig:geometry_regime} with precise numerical values.
Figure~\ref{fig:spider} then extends the analysis to the category level,
comparing the full model (FM), the model without the velocity pretext (NVP) and the model without BN (NBN) across the 17 dataset families in TSB-AD-M, and
showing how consistently the geometric separation observed in aggregate arises
at the individual category level.

\begin{table}[t]
  \caption{Embedding geometry diagnostics (mean $\pm$ std over 10 seeds,
           180-series held-out set).}
  \label{tab:geometry}
  \centering
  \setlength{\tabcolsep}{6pt}
  \resizebox{\textwidth}{!}{
  \begin{tabular}{l cccc}
    \toprule
    Variant
      & \makecell{Eff.\ rank\\($\rho_{\mathrm{PR}}$/$d$)}
      & \makecell{Eff.\ rank\\($\rho_H$/$d$)}
      & \makecell{Active\\dims}
      & \makecell{Top-1\\var.\ frac.} \\
    \midrule
    w/o Velocity Pretext  & $0.081\pm0.004$ & $0.143\pm0.005$ & $0.953\pm0.006$ & $0.493\pm0.015$ \\
    w/o BN       & $0.025\pm0.001$ & $0.037\pm0.002$ & $0.853\pm0.021$ & $0.816\pm0.014$ \\
    \midrule
    \rowcolor{gray!12}
    Full model            & $0.061\pm0.002$ & $0.115\pm0.002$ & $0.996\pm0.001$ & $0.551\pm0.010$ \\
    \bottomrule
  \end{tabular}
  }
\end{table}

\paragraph{Velocity pretext.}
The pretext task reduces effective rank in 120 of 180 series (67\%) and increases $n_{\mathrm{active}}$ in 152 of 180 (84\%), from 61.0 to 63.7 active dimensions. These effects are consistent. The objective sharpens dominant principal components and lifts near-zero dimensions above the activity threshold. As a result, the eigenspectrum shifts from near-uniform to a spike-and-tail structure. \revadd{The resulting geometry therefore offers} clear high-variance directions for the Mahalanobis distance to exploit, and a well-populated embedding space with no wasted capacity. Twenty gradient steps of velocity consistency produce this structure reliably across 16 of 17 datasets.

\paragraph{Channel encoder.}
The channel encoder's effect on geometry is heterogeneous and scales with the number of input channels. For series with two to five channels, the channel encoder leaves the effective rank essentially unchanged ($\Delta\rho_{\mathrm{PR}}/d = {+0.003}$, near zero), and slightly reduces the top-1 variance fraction ($-0.026$).  This suggests that with very few channels, the depthwise stage has limited per-channel structure to extract before mixing, and the encoder converges to a solution similar to shared-kernel processing.

For series with more than 20 channels, the channel encoder consistently reduces effective rank and concentrates variance. The effect is strongest for the highest-channel group (71--248 channels, $n = 7$, $\Delta$top-1 $= {+0.124}$), where early cross-channel mixing in the shared-kernel baseline dilutes per-sensor patterns across many channels before any discriminative feature can form. The MSL category, as shown in Table \ref{tab:msl_geometry}, illustrates this division: MSL (55 channels) shows a large increase in active dimensions under the channel encoder ($n_{\mathrm{active}}$ increases from $48.9$ to $64.0$) with respect to the no channel-aware encoder ablation, indicating that the shared-kernel baseline leaves ${\approx}15$ embedding dimensions inactive.

\begin{table}[h]
  \caption{Per-series geometry and detection performance on MSL (14 evaluation series, mean over 10 seeds). Sorted by $\Delta$VUS-PR. $n_{\mathrm{active}}$ increases in all 14 series. VUS-PR improves in 8 of 14. NCA: model without the channel-aware encoder. FM: full model.}
  \label{tab:msl_geometry}
  \centering
  \setlength{\tabcolsep}{6pt}
  \begin{tabular}{l | cc r | cc}
    \toprule
    & \multicolumn{3}{c|}{VUS-PR} & \multicolumn{2}{c}{$n_{\mathrm{active}}$ / 64} \\
    \cmidrule(lr){2-4}\cmidrule(lr){5-6}
    Series & NCA & FM & $\Delta$ & NCA & FM \\
    \midrule
    MSL id\_1  & 0.087 & \textbf{0.665} & $+$0.578 & 47 & \textbf{64} \\
    MSL id\_13 & 0.485 & \textbf{0.792} & $+$0.308 & 46 & \textbf{64} \\
    MSL id\_12 & 0.482 & \textbf{0.713} & $+$0.231 & 47 & \textbf{64} \\
    MSL id\_14 & 0.567 & \textbf{0.766} & $+$0.198 & 48 & \textbf{64} \\
    MSL id\_15 & 0.490 & \textbf{0.530} & $+$0.041 & 45 & \textbf{64} \\
    MSL id\_4  & 0.262 & \textbf{0.294} & $+$0.032 & 52 & \textbf{64} \\
    MSL id\_11 & 0.095 & \textbf{0.122} & $+$0.027 & 48 & \textbf{64} \\
    MSL id\_2  & 0.107 & \textbf{0.117} & $+$0.010 & 52 & \textbf{64} \\
    \midrule
    MSL id\_6  & 0.071 & 0.071 & $\phantom{+}$0.000 & 52 & \textbf{64} \\
    MSL id\_9  & 0.068 & 0.043 & $-$0.025 & 48 & \textbf{64} \\
    MSL id\_7  & 0.441 & 0.384 & $-$0.057 & 47 & \textbf{64} \\
    MSL id\_5  & 0.142 & 0.075 & $-$0.067 & 51 & \textbf{64} \\
    MSL id\_16 & 0.302 & 0.234 & $-$0.068 & 48 & \textbf{64} \\
    MSL id\_8  & 0.171 & 0.012 & $-$0.159 & 51 & \textbf{64} \\
    \bottomrule
  \end{tabular}
\end{table}

\paragraph{Batch normalisation.}
Removing BN produces a qualitatively different failure than removing the pretext. Effective rank falls to $0.025$ across all 17 categories ($0.017$--$0.046$), and $n_{\mathrm{active}}$ drops to $54.6$ out of 64: roughly nine dimensions become inactive. The no-pretext variant also loses geometric structure but retains full dimensional occupancy ($n_{\mathrm{active}} = 61.0$). Collapse compounds two failures. On the one hand, the spectrum becomes concentrated but uninformative. On the other hand, the covariance matrix becomes nearly singular, and its inverse is dominated by near zero eigenvalues rather than meaningful directions.

\section{Limitations}
\label{app:limitations}

Three structural limitations bound the applicability of VACE. First, very long training series produce a large number of overlapping patches, increasing the volume of the embedding point cloud and making the single-Gaussian
Mahalanobis fit a noisier approximation of the normal region. Second, the channel-aware encoder processes channels independently before mixing, which is beneficial when anomalies manifest in individual sensors but may discard discriminative inter-channel correlation structure when anomalies are defined by unusual relationships between jointly behaving channels. Third, reconstruction-based methods outperform VACE on a subset of categories where the geometric scoring signal is insufficient. \revadd{Appendix~\ref{sec:additional_experiments} identifies two properties that account for part of this gap, normal regions without a dominant direction and level-coded anomalies that the per-patch normalisation of RevIN hides, but the correlations involved are moderate and the explanation is not complete.}

\begin{revaddblock}
Further limitations apply. (i) The training objective applies no explicit pressure separating anomalies from normal data. The premise that anomalies land away from the normal trajectory holds empirically on most series, but fails on GHL, where anomalies lie in the centre of the normal region (Appendix~\ref{app:premise}). (ii) The benchmark average is carried mostly by the positional score. The directional score adds a small, systematic gain and is strongly informative only on some categories (Table~\ref{tab:combination}). (iii) The direction bank is an unordered set of normal directions with no time index, so a direction that is normal in itself but appears at the wrong time is not flagged by the directional score. Preserving temporal order in the directional score is left as future work. (iv) The positional score models the normal region with a single Gaussian, which is the weak assumption on the series where VACE fails. A richer density over the same representation, from a mixture to the latent-variable models used by ASTER \cite{hermary2026aster}, could capture the non-stationarity and the multiple operating regimes that a single Gaussian cannot, while still requiring no model of the anomalies. (v) VACE as evaluated is an offline detector, and the score of a timestep needs the next $P - 1 + \delta$ observations (Section~\ref{sec:scoring}). A causal variant has not been evaluated. (vi) The velocity-consistency objective has a degenerate, straight-line minimiser. The deployed configuration avoids it through its short schedule, decaying loss weight and BN, but the objective itself does not prevent it (Appendix~\ref{app:collapse}).
\end{revaddblock}

\section{Broader Impacts}
\label{app:broader_impacts}

VACE is an anomaly detection method for multivariate time series, with potential positive applications in industrial monitoring, server infrastructure management,
and medical signal analysis, where detection of abnormal behaviour can prevent failures or inform clinical decisions. The method operates on anomaly-free training data and produces anomaly scores without any decision-making authority. Deployment decisions remain under human control. We identify no direct path to negative societal applications.

\end{document}